\documentclass[lettersize,journal]{IEEEtran}
\usepackage{amsmath,amsfonts}
\usepackage{algorithmic}
\usepackage{array}
\usepackage{textcomp}
\usepackage{stfloats}
\usepackage{url}
\usepackage{verbatim}
\usepackage{graphicx}
\usepackage{algorithm}
\usepackage{booktabs}
\usepackage{xcolor}
\usepackage{caption}
\usepackage{subcaption}
\usepackage[numbers]{natbib}

\def\BibTeX{{\rm B\kern-.05em{\sc i\kern-.025em b}\kern-.08em
    T\kern-.1667em\lower.7ex\hbox{E}\kern-.125emX}}
\usepackage{balance}
\begin{document}
\title{Inferring Latent Temporal Sparse Coordination Graph for Multi-Agent Reinforcement Learning}

\author{Wei~Duan,~\IEEEmembership{Student Member,~IEEE,} ~Jie~Lu,~\IEEEmembership{Fellow,~IEEE, }
and Junyu~Xuan,~\IEEEmembership{Senior Member,~IEEE}% <-this % stops a space
\thanks{The authors are with the Australian Artificial Intelligence Institute (AAII), Faculty of Engineering and Information Technology, University of Technology Sydney, Ultimo, NSW 2007, Australia. (Email:
wei.duan@student.uts.edu.au, jie.lu@uts.edu.au,junyu.xuan@uts.edu.au)
}
}

\markboth{Journal of \LaTeX\ Class Files,~Vol.~X, No.~X, 2024}%
{How to Use the IEEEtran \LaTeX \ Templates}

\maketitle

\begin{abstract}
Effective agent coordination is crucial in cooperative Multi-Agent Reinforcement Learning (MARL). While agent cooperation can be represented by graph structures, prevailing graph learning methods in MARL are limited. They rely solely on one-step observations, neglecting crucial historical experiences, leading to deficient graphs that foster redundant or detrimental information exchanges. 
Additionally, high computational demands for action-pair calculations in dense graphs impede scalability. To address these challenges,  we propose inferring a Latent Temporal Sparse Coordination Graph (LTS-CG) for MARL. The LTS-CG leverages agents' historical observations to calculate an agent-pair probability matrix, where a sparse graph is sampled from and used for knowledge exchange between agents, thereby simultaneously capturing agent dependencies and relation uncertainty. The computational complexity of this procedure is only related to the number of agents. This graph learning process is further augmented by two innovative characteristics: Predict-Future, which enables agents to foresee upcoming observations, and Infer-Present, ensuring a thorough grasp of the environmental context from limited data. These features allow LTS-CG to construct temporal graphs from historical and real-time information, promoting knowledge exchange during policy learning and effective collaboration. Graph learning and agent training occur simultaneously in an end-to-end manner. Our demonstrated results on the StarCraft II benchmark underscore LTS-CG's superior performance.
\end{abstract}


\section{Introduction}
Effective agent coordination is crucial in cooperative Multi-Agent Reinforcement Learning (MARL), which offers an instrumental approach to control multiple intelligent agents to fulfil various tasks, including coordinating traffic lights throughout a city \cite{DBLP:journals/tits/0017W0H22}, orchestrating multi-robot formations \cite{DBLP:journals/csur/RizkAT19}, and optimizing the behaviour of unmanned aerial vehicles \cite{DBLP:journals/twc/CuiLN20} 
One efficient approach to training multiple agents in dynamic environments involves decomposing the global value function into manageable segments for each agent. This methodology is exemplified by techniques such as VDN employing the sum of independent agent value functions \cite{DBLP:conf/atal/SunehagLGCZJLSL18}, QMIX utilizing a monotonic mixture instead of a simple sum \cite{DBLP:conf/icml/RashidSWFFW18}, and QTRAN using a hyper-edge that connects all agents without factorization \cite{DBLP:conf/icml/SonKKHY19}. Within this framework, each agent selects actions to maximize its own value function and contributes to maximising the total reward.

\begin{figure}[t]
\centering
\includegraphics[width=\columnwidth]{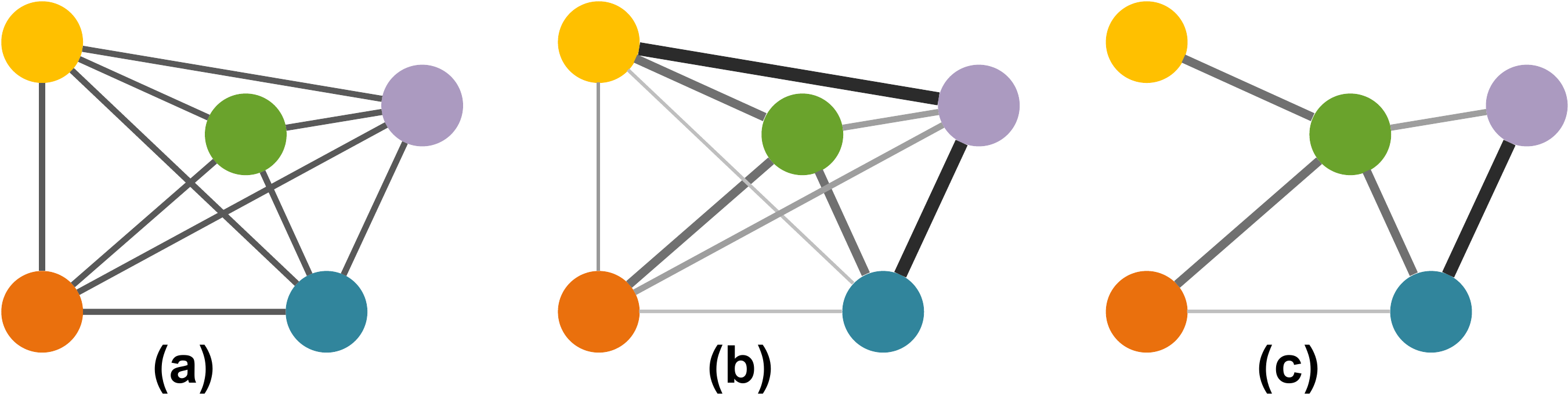} % Reduce the figure size so that it is slightly narrower than the column. Don't use precise values for figure width.This setup will avoid overfull boxes.
\caption{The current methods to infer latent graphs in MARL can be categorized into three types: (a) fully connected unweighted graphs, (b) fully connected weighted graphs, and (c) sparse weighted graphs. These methods rely solely on one-step observations, leading to deficient graphs that foster redundant or detrimental information exchanges and suffer from high computational complexity for action-pair calculations.} %The current methods suffer from redundant information exchange in fully-connected graphs, overlooking crucial historical trajectory data during graph learning and incurring high computational complexity for action-pair calculations.}
\label{fig:3stageGMARL}
\end{figure}

While these methods balance computational efficiency with effective agent interaction and complex decision-making, in the real world, agents should not only consider their own observations but also take into account the situations of others when taking action \cite{DBLP:conf/nips/HongJT22}. Effective cooperation among agents emerges as a pivotal factor in achieving specific objectives.
This cooperation can be assumed to have some latent graph structures \cite{DBLP:conf/icml/GuestrinLP02}. Since the \textcolor{black}{agent graph} is not explicitly given, the inference of meaningful dynamic graph topology has been a persistent challenge. 

The current methods to address this problem can be broadly categorized into three types, illustrated in Fig.\ref{fig:3stageGMARL}. The first type involves employing fully connected unweighted graphs, such as PIC \cite{pmlr-v100-liu20a} and DCG \cite{DBLP:conf/icml/BoehmerKW20}. The second type incorporates fully connected weighted graphs, such as  GraphMIX \cite{DBLP:journals/corr/abs-2010-04740} and DICG\cite{DBLP:conf/atal/LiGMAK21}. The third type utilizes weighted sparse graphs, such as SOP-CG \cite{DBLP:conf/icml/YangDRW0Z22} and CASEC \cite{,DBLP:conf/iclr/00010DY0Z22}. However, these methods exhibit the following limitations: (1) They primarily focus on one-step observations and fail to consider the value of historical trajectory data, which more accurately represents agents' behaviours and is more meaningful to help to learn policies \cite{pmlr-v119-pacchiano20a}. This overreliance on one-step data can lead to suboptimal graph learning, producing graphs that may encourage redundant or even counterproductive information exchanges, thereby impeding effective policy learning. (2) The computation-intensive nature of action-pair calculations in coordination graphs (CG) \cite{DBLP:conf/icml/GuestrinLP02} poses significant scalability challenges, especially in fully-connected settings. For instance, in a system with $N$ agents, each having $A$ actions, the computational complexity of these methods is $O(A^2N^2)$. This complexity becomes increasingly problematic as the number of agents and actions increases.
%In these two types of methods, adopting a fully connected graph inevitably transfers redundant or even detrimental information between agents when not all agents necessitate it for decision-making. This could hinder their ability to learn effective policies. Despite the third type employing weighted sparse graphs \cite{DBLP:conf/icml/YangDRW0Z22,DBLP:conf/iclr/00010DY0Z22}, subsequent experiments reveal that these methods require high computational demands for action-pair calculations, posing scalability challenges with a larger agent scenario. Moreover, the graph learning process of all the above methods is limited to the one-step observation, overlooking potentially valuable historical trajectory data and forgetting past experience for cooperation. 

In this paper, we address these limitations by proposing a novel approach called Latent Temporal Sparse Coordination Graph (LTS-CG) for MARL. LTS-CG efficiently infers graphs using agents' observation trajectories to generate an agent-pair probability matrix, where the probability is absorbed and trained together with Graph Convolutional Networks (GNN) parameters. The computational complexity of this procedure scales quadratically with the number of agents $N$, which renders our approach scalable and suitable for handling complex MARL scenarios. Subsequently, a sparse graph is sampled from this matrix, which simultaneously captures agent dependencies underlying the trajectories and models the relation-uncertainty between agents. Driven by the goal of creating meaningful graphs, we enhance agents' understanding of their peers and the environment by embedding two essential characteristics into the graph: Predict-Future and Infer-Present.  Predict-Future empowers agents to predict upcoming observations using current observations and the sampled graph, providing valuable insights for immediate decision-making. Infer-Present aids each partially observed agent in comprehending the full environmental context and deducing the current state with the graph's information. LTS-CG leverages both historical and real-time data for graph training, considering local and global perspectives. The temporal structure of the learned graph encapsulates past experiences, with edge weights reflecting ongoing observations. This facilitates knowledge exchange during policy learning and supports historical and present insights for effective cooperation. The computational complexity of our method is $O(TN^2)$, where $T$ represents the observation length used for graph learning, making it more efficient than action-pair-based methods.

The main insight behind designing our method is to enable simultaneous graph inference and multi-agent policy learning, facilitating efficient end-to-end training using standard policy optimization methods. We evaluate LTS-CG on the StarCraft II benchmark, demonstrating its superior performance. The ablation results empirically proved that using trajectories for learning the coordination graph is more effective than relying on one-step observations, and having the Predict-Future and Infer-Present characteristics improves the performance of LTS-CG. The contributions of this paper are summarized as follows:
\begin{itemize}
    \item  We pioneer the treatment of agent trajectories as data streams in MARL with LTS-CG. Our method leverages these trajectories to infer latent temporal sparse graphs, facilitating knowledge exchange between agents.
    \item By sampling sparse graphs from trajectories-generated agent probability matrices, LTS-CG captures agent dependencies and models the uncertainty of relations between agents simultaneously, with computational complexity only related to the number of agents.
    \item LTS-CG further infers the graph from both local and global standpoints to encode Predict-Future and Infer-Present characteristics. This meaningful graph enables agents to gain historical and present perspectives to achieve effective cooperation.
\end{itemize}

The rest of the paper is organized as follows. In Sec.~\ref{sec:pre}, we give a definition of our task, followed the related work in Sec.~\ref{sec:related work}. In Sec. \ref{sec:method}, we described our approach. We report experimental studies
in Sec. \ref{sec:exp} and conclude in Sec. \ref{sec:conclusion}.

\begin{figure*}[h]
\centering
\includegraphics[width=\textwidth]{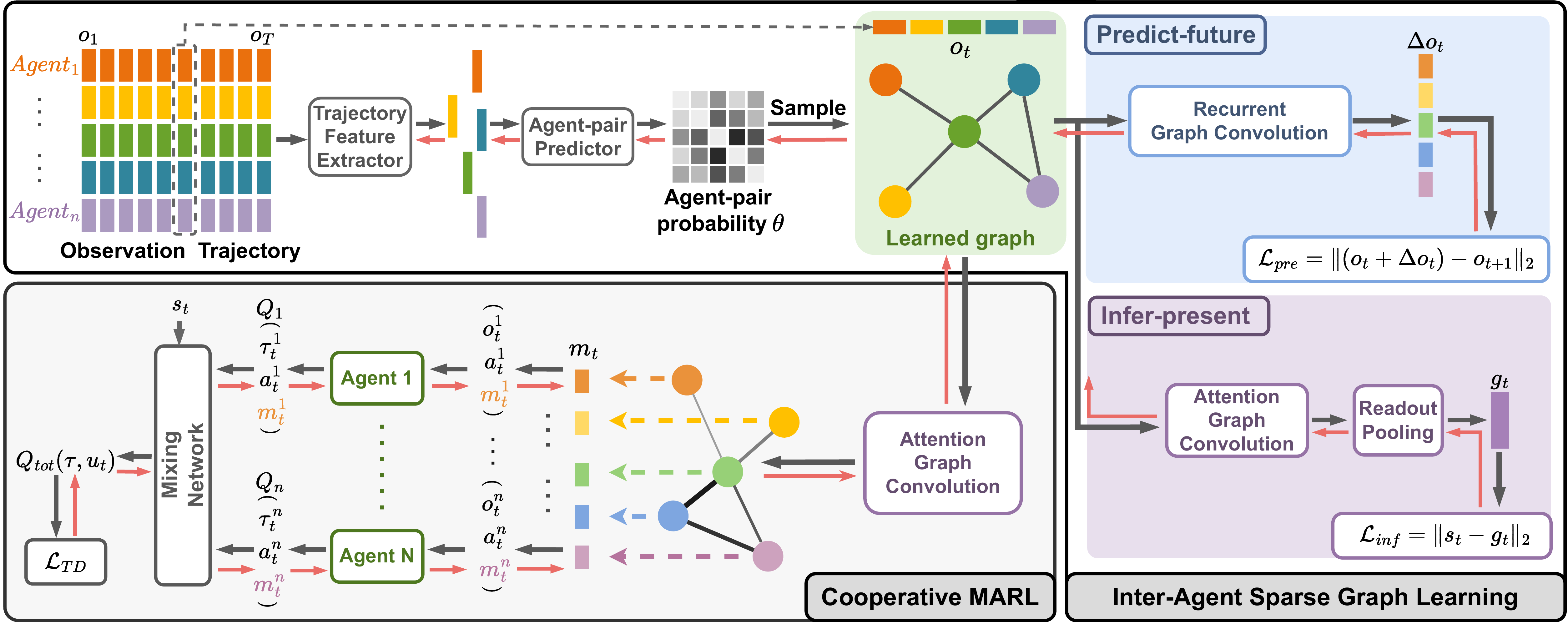} % Reduce the figure size so that it is slightly narrower than the column
\caption{The framework of LTS-CG. \textcolor{black}{ LTS-CG consists of two key modules: \textbf{Inter-Agent Sparse Graph Learning} and \textbf{Cooperative MARL}. The former follows an encoder-decoder framework: the encoder generates the sparse graph structure, while the decoder—guided by two graph loss functions—learns Predict-Future for anticipating future steps and Infer-Present for deducing current states. The temporal graph structure integrates past experiences and adjusts edge weights based on current observations. This graph is then fed into the attention-based graph convolution of the \textbf{Cooperative MARL} module, enabling knowledge exchange for effective coordination. Graph learning and agent training occur end-to-end.}}
\label{fig:framework}
\end{figure*}

\section{Preliminaries}
\label{sec:pre}
We focus on cooperative multi-agent tasks modelled as a \textcolor{black}{Partially Observable Markov Decision Process (POMDP) \cite{du2021survey}} consisting of a tuple $\langle \mathcal{I}, \mathcal{S},\{\mathcal{A}^{i}\}_{i=1}^{n}, P,\{\mathcal{O}^{i}\}_{i=1}^{n},\{\sigma^{i}\}_{i=1}^{n}, R,\gamma\rangle$, where $\mathcal{I}$ is the finite set of $n$ agents, $s \in \mathcal{S}$ is the true state of the environment. At each time step, each agent $i$ observes the state partially by drawing observation $o^{i}_{t} \in \mathcal{O}^{i}$ and selects an action $a^{i}_{t} \in \mathcal{A}^{i}$ according to its own policy $\sigma^{i}$. Individual actions form a joint action $\boldsymbol{a}=(a_1,...,a_n)$, which leads to the next state $s^{\prime}$ according to the transition function $P(s^{\prime}| s, \boldsymbol{a})$ and a reward $R(s,\boldsymbol{a})$ shared by all agents. Each agent has local action-observation history $\tau_{i,t}=(o_{i,0},a_{i,0},...,o_{i,t-1},a_{i,t-1},o_{i,t}) \in (\mathcal{O}^{i} \times \mathcal{A}^i)^{t}\times \mathcal{O}^{i}$. This paper considers episodic tasks yielding episodes $(s_0,\{o^{i}_{0}\}_{i=1}^{n},\boldsymbol{a}_0,r_{0},...,s_T,\{o^{i}_{T}\}_{i=1}^{n})$ of varying finite length $T$. Agents learn to collectively maximize the global return $Q_{tot}(s, \boldsymbol{a})=\mathbb{E}_{s_{0: T}, a_{0: T}}\left[\sum_{t=0}^{T} \gamma^t R\left(s_t, \boldsymbol{a}_t\right) \mid s_0=s, \boldsymbol{a}_0=\boldsymbol{a}\right]$, where $\gamma \in [0,1)$ is the discount factor.

Learning the underlying relation of agents can be seen as the inference of a meaningful dynamic graph topology. This graph is denoted as $\mathcal{G}=\{\mathcal{V},\mathcal{E}\}$ where $\mathcal{V}=\mathcal{I}$ is node/agent set and $\mathcal{E}$ is the edge/relation set between agents. 
 
\section{Related Work}
\label{sec:related work}
\subsection{Graph-based MARL}
MARL faces the challenge of dealing with the exponentially growing size of joint action spaces among agents \cite{DBLP:journals/apin/OroojlooyH23}. The paradigm of CTDE \cite{DBLP:conf/nips/LoweWTHAM17,DBLP:conf/aaai/FoersterFANW18} strikes a balance between computational efficiency and multi-agent interaction but falls short in handling dependencies between agents. Graph Neural Networks (GNNs) \cite{DBLP:conf/iske/DuanX021,DBLP:journals/eswa/YaoHLDQYS25} have demonstrated remarkable capability in modelling relational dependencies \cite{DBLP:journals/tnn/WuPCLZY21,DBLP:conf/aaai/LiuWHHC020}, making graphs a compelling tool for graph-based MARL, which can be generally divided into two types. 
One type involves using graphs as coordination graphs during policy training, such as DCG \cite{DBLP:conf/icml/BoehmerKW20}, SOP-CG \cite{DBLP:conf/icml/YangDRW0Z22} and CASEC \cite{,DBLP:conf/iclr/00010DY0Z22}. In this approach, the total action-value function is defined as:
\begin{equation}
    Q_{tot}\!\left(s_t, \boldsymbol{a}\right)\!=\!\frac{1}{|\mathcal{V}|}  \sum_{i \in \mathcal{V}} q^i\left(a^i \! \mid \! s_t\right)\!+\!\frac{1}{|\mathcal{E}|}\! \sum_{\{i, j\} \in \mathcal{E}}\!\!\! q^{ij}\!\left(a^i, a^j \! \mid \! s_t\right),
\label{eq:CG}
\end{equation}
where the first term calculates the Q-value of each action (also known as utility function), and the second term evaluates every action-pair of agents (also known as payoff function). This method explicitly assesses the quality of joint actions between different agents.
The other type uses graphs to facilitate information exchange among agents, such as DICG \cite{DBLP:conf/iclr/0001WZZ20} and G2ANet \cite{DBLP:conf/aaai/DuanXQ022}. It is formulated as:
\begin{equation}
m_{i}= \text{AGG}_{j \in \mathcal{N}_i}(f(o_{j},a_j)), \quad Q_{tot} = \sum_{i=1}^{n}Q_i(o_{i},a_i,m_{i})
\end{equation}
where $\mathcal{N}_i$ means the neighbours of agent $i$. $f(\cdot)$ transfers the original observation and action into embedding, and AGG($\cdot$) aggregates the embedding based on graph topology to generate the message $m_{i}$. This message provides additional knowledge that aids agents in decision-making and represents an implicit coordination between agents. Although these methods do not strictly calculate the payoff-utility function based on the coordination graph, they build upon the same idea of reasoning about joint actions based on interactions between agents \cite{DBLP:conf/iclr/0001WZZ20}.

As the graph itself is not explicitly given, inferring graph topology remains a critical prerequisite for training MARL. From the perspective of graph structure, existing methods for graph inference can be broadly categorized into three types: (a) creating fully connected unweighted graphs by directly linking all nodes/agents explicitly, such as DGN \cite{DBLP:conf/iclr/JiangDHL20}, PIC \cite{pmlr-v100-liu20a} and DCG \cite{DBLP:conf/icml/BoehmerKW20}, or implicitly such as MAAC \cite{DBLP:conf/icml/IqbalS19}, ROMA \cite{DBLP:conf/icml/0001DLZ20}; (b) employing attention mechanisms to calculate fully connected weighted graphs, such as GraphMIX \cite{DBLP:journals/corr/abs-2010-04740} and DICG \cite{DBLP:conf/atal/LiGMAK21}; (c) designing drop-edge criteria to generate sparse weighted graphs, such as random drop edges in G2ANet \cite{DBLP:conf/aaai/LiuWHHC020}, select sparse graph from candidate set in SOP-CG \cite{DBLP:conf/icml/YangDRW0Z22}, drop edges based on variance of payoff functions in CASEC \cite{DBLP:conf/iclr/00010DY0Z22}, and and \textcolor{black}{generate event graphs with certain rules in CAAC \cite{DBLP:conf/ijcai/WangS21}}.

Despite this progress, these methods exhibit the following limitations: one is that they primarily focus on one-step observations and fail to consider the value of historical trajectory data, which more accurately represents agents' behaviours and is more meaningful to help to learn policies \cite{pmlr-v119-pacchiano20a}; another is that the computation-intensive nature of action-pair calculations in coordination graphs (CG) \cite{DBLP:conf/icml/GuestrinLP02} poses significant scalability challenges, which becomes increasingly problematic as the number of agents and actions increases (See: \ref{sec:Graph-based} and \ref{sec:Dsicussion}).

\subsection{Graph Structure Learning}
To learn a relational graph between agents that take a series of actions within specific time steps, two promising directions are worth considering: learning a graph for multiple time series forecasting and inferring a graph for trajectory prediction. For the former, \citet{DBLP:conf/ijcai/YuYZ18} explored pairwise similarities or connections among them to enhance forecasting accuracy. \citet{DBLP:conf/kdd/WuPL0CZ20} presented a framework for modelling multivariate time series data and learning graph structures that can be used with or without a pre-defined graph structure. \citet{DBLP:journals/corr/abs-2203-03423} proposed an approach that balances accuracy and computational efficiency, allowing the flexibility to infer either fully connected or bipartite graphs.
Regarding trajectory prediction, \citet{DBLP:conf/icml/KipfFWWZ18} proposed NRI, a variational autoencoder that leverages a latent-variable approach to learn a latent graph. On the other hand, LDS \cite{DBLP:conf/icml/FranceschiNPH19} and GTS \cite{DBLP:conf/iclr/Shang0B21} focus on learning probabilistic graph models by optimizing performance over the graph distribution mean. To further adaptively connect multiple nodes, \citet{DBLP:journals/corr/abs-2208-05470} proposed a group-aware relational reasoning approach to infer hyperedges. In the context of MARL, the absence of labelled data poses a challenge for traditional trajectory prediction or multiple time series forecasting methods. Borrowing the learning capabilities from these two directions while fully leveraging the information available in MARL remains an underexplored area.

\section{The Proposed Method}
\label{sec:method}
The framework of LTS-CG is illustrated in Fig. \ref{fig:framework}.
To efficiently infer the underlying relation from past experiences, LTS-CG samples a sparse graph from the agent-pair probability matrix generated by agents' observation trajectories. The core of LTS-CG lies in creating a meaningful graph that enhances agents' understanding of their peers and the environment. This is achieved through two key characteristics: Predict-Future and Infer-Present, which enable agents to share knowledge and gain both historical and present insights, fostering effective cooperation. Detailed descriptions of each component are provided in the subsequent sections.

\subsection{Latent Temporal
Sparse Graphs Learning}
\subsubsection{Sparse Graph Construction}
The accumulated observation trajectories of all agents encapsulate their experiences of interactions with the environment and their cooperation. To efficiently capture the underlying relationships, instead of directly learning the structure of the inter-agent sparse graph $A$, we utilize observation trajectories $\{\mathcal{O}^{i}\}_{i=1}^{n}$ to generate the agent-pair probability matrix $\theta \in [0,1]^{n \times n}$. This matrix parameterizes the element-wise Bernoulli distribution \cite{DBLP:conf/icml/FranceschiNPH19}, which allows us to sample a graph representing the relevant connections between agents. This graph learning objective is achieved by minimizing the loss of function
\begin{equation}
\textcolor{black}{
\min _\omega \quad \mathrm{E}_{A \sim \operatorname{Ber}(\theta(\omega))}\left[\mathcal{L}\left(A, \omega, \mathcal{O}_T \right)\right] .}
\label{eq:learnMatrix}
\end{equation}
Here, $\mathcal{O}_T =\{{\mathcal{O}^{i}_{T}\}_{i=1}^{n}}$, and $\mathcal{O}^{i}_{T}=\{o^{i}_{0},...,o^{i}_{T}\}$ denotes the observation trajectory for agent $i$ over the time steps $T$. Each element of $A$ is sampled from a Bernoulli distribution \textcolor{black}{$\operatorname{Ber}(\theta(\omega))$, with $\omega$} denoting the trainable weight. In Eq. (\ref{eq:learnMatrix}), the adjacent probability $\theta$ is absorbed together with the GNNs parameters \textcolor{black}{$\omega$}, making the gradient computation more efficient and having better scalability \cite{DBLP:conf/iclr/Shang0B21}.  In the following, we give the details about how to infer the inter-agent sparse graph $A$ and how to define the graph learning loss function  $\mathcal{L}$.

To acquire knowledge about the temporal dependence of each agent and the relationship between agents, we establish the
observation experience extractor $f_{oe}(\cdot)$ to help us capture the temporal dependence of each agent $z^i$ by employing convolution along the time dimension, followed by a fully connected layer, defined as
\begin{equation}
    z^i = f_{oe}(\mathcal{O}^{i}_{T})
        = \text{FC}(\text{CONV}(\mathcal{O}^{i}_{T})),
\end{equation}
where $\text{FC}(\cdot)$ is a fully connected layer and $\text{CONV}(\cdot)$ is the convolution layer performed along the temporal dimension. \textcolor{black}{Since the episodes may end before reaching the maximum time step, zeros are padded in $\mathcal{O}^{i}_{T}$ to ensure consistent input length across episodes.} This convolutional layer plays a crucial role in capturing each agent's latent behaviour patterns over time, enhancing the model's ability to discern dynamic and temporal patterns in the agents' interactions. Then the agent-pair predictor $f_{ap}(\cdot)$ utilize the temporal dependencies of every agent-pair ($z^i$ and $z^j$) to calculate adjacent probability $\theta_{ij}$ as follows
\begin{equation}
\theta_{ij} = f_{ap}(z^i \| z^j)  
                = \text{FC}(\text{FC}(z^i \| z^j)),
\label{eq:obs-emb}
\end{equation}
where $\|$ denotes concatenation along the feature dimension. We adopt multi-layer perceptrons (MLPs) to model and learn $f_{ap}(\cdot)$, leveraging the universal approximation theorem \cite{DBLP:journals/nn/Hornik91} to enhance their representational capacity. 

To enable backpropagation through the Bernoulli sampling, we apply the Gumbel parameterization trick \cite{DBLP:conf/iclr/JangGP17,DBLP:conf/iclr/MaddisonMT17}. This technique leverages the properties of the Gumbel distribution to approximate the sampling process in a differentiable manner, allowing gradients to flow through the stochastic operation. In the context of Bernoulli sampling, the Gumbel trick involves generating two Gumbel-distributed random variables, denoted as $g_{ij}^1$ and $g_{ij}^2$, for each element $A_{ij}$ in the adjacency matrix. These random variables are sampled from a Gumbel distribution with a location parameter of 0 and a scale parameter of 1.
The sampled values from the Gumbel distribution are then used to compute the logits for the sigmoid function in the Bernoulli sampling equation. Specifically, the logits are calculated as:
\begin{equation}
A_{i j} \!=\operatorname{sigmoid}\left(\left(\log \left(\theta_{i j} /\!\left(1\!-\!\theta_{i j}\right)\right)\!+\!\left(g_{i j}^1\!-\!g_{i j}^2\right)\right)\! / s\right),
\label{ed:adj-sample}
\end{equation}
where $g_{i j}^1, g_{i j}^2 \sim \operatorname{Gumbel}(0,1)$ for all $i, j$, $\theta_{ij}$ represents the probability parameter for the Bernoulli distribution, and $s$ is a temperature parameter that controls the sharpness of the sampling process. As the temperature $s \rightarrow 0, A_{i j}=1$ with probability $\theta_{i j}$ and 0 with remaining probability. By applying Eq. (\ref{eq:obs-emb}) and Eq. (\ref{ed:adj-sample}), we convert the observation trajectories $\mathcal{O}_T$ into an agent-pair probability $\theta$. We subsequently sample to obtain the inter-agent graph $A$ for further learning and utilization in cooperative MARL.

\subsubsection{Meaningful Graph Learning}
Motivated by the idea that the graph should enhance the agents' understanding of other agents and the environment, we further learn the graph to have the following two essential characteristics.

% In contrast to supervised learning, lending the learning capabilities from traditional trajectory prediction or multiple time series forecasting methods to MARL encounters the challenge of lacking readily available labelled data for graph learning. To overcome this obstacle, we exploit the rich environmental information inherent in MARL to refine and train the inter-agent graph $A$. Our motivation is two-fold: (1) by utilizing the graph, we aim to empower agents to predict future steps effectively, enabling them to make better decisions in the current time step; %This predictive capability enhances the overall performance of the cooperative MARL. 
% (2) considering that each agent only has access to partial observations, we leverage graph convolution to enable agents to seamlessly exchange their observations, allowing the entire graph (all agents and their relationships) to represent the current state. In light of this, we introduce the \textbf{Predict-Future Decoder} and the \textbf{Infer-Present Decoder} components.
% Through these decoders, LTS-CG maximizes the utilization of available 
% historical and current environmental information to train the graph from both local and global perspectives. Further, the learned graph is then used for information exchange between agents to help each agent learn policy and facilitate cooperation.

\noindent \textbf{Predict-Future} means by exploiting the graph, we aim to empower agents to predict future steps effectively, enabling them to make better decisions in the current time step. We use the diffusion convolutional gated recurrent unit introduced in Diffusion Convolutional Recurrent Neural Network (DCRNN) \cite{DBLP:conf/iclr/LiYS018} and leverage the learned graphs $A$ to process the observations of all agents $\mathcal{O}_{t}={\{o^{i}_t\}_{i=1}^{n}}$ as follows
\begin{equation}
\begin{array}{ll}
& R_{t}=\operatorname{sigmoid}\left(W_R \star_A\left[\mathcal{O}_{t} \| H_{t-1}\right]+b_R\right), \\
& C_{t}=\tanh \left(W_C \star_A\left[\mathcal{O}_{t}\|\left(R_{t} \odot H_{t-1}\right]+b_C\right)\right. \\
& U_{t}=\operatorname{sigmoid}\left(W_U \star_A\left[\mathcal{O}_{t}\| H_{t-1}\right]+b_U\right), \\
& H_{t}=U_{t} \odot H_{t-1}+\left(1-U_{t}\right) \odot C_{t},
\end{array}    
\label{eq:dcrnn}
\end{equation}
where the graph convolution $\star_A$ is defined as
\begin{equation}
W_Q \star_A Y=\sum_{k=0}^K\left(w_{k, 1}^Q\left(D_O^{-1} A\right)^k+w_{k, 2}^Q\left(D_I^{-1} A^T\right)^k\right) Y,
\end{equation}
with $D_O$ and $D_I$ being the out-degree and in-degree matrix of learned agent-pair matrix $A$, respectively. Here, $w_{k, 1}^Q, w_{k, 2}^Q, b_Q$ for $Q=R, U, C$ are model parameters and $K$ is the diffusion degree. We adopt a $1$-layer DCRNN and set $K=3$ in our experiments.

To capture both temporal and spatial dependencies between agents, we feed a $T$-step observations ${\{o^{i}_{t+1:t+T}\}_{i=1}^{n}}$ into Eq.(\ref{eq:dcrnn}), to forecast the future changes in the current $T$-step observation. The output of the hidden state in every step represents the prediction of how the current observation will change in the next step, denoted as  $H_{t+1:t+T}={\{\Delta o^{i}_{t+1:t+T}\}_{i=1}^{n}}$. Then, the Predict-Future is achieved by calculating the following loss function
\begin{equation}
\mathcal{L}_{pre}= \sum_{i}^{n} \sum_{t^{\prime}=1}^{T}\left\| \left( o^{i}_{t+t^{\prime}}+ \Delta o^{i}_{t+t^{\prime}} \right) - o^{i}_{t+1+t^{\prime}}\right\|_{2}.
\label{eq:l-pre}
\end{equation}
Since Eq.~(\ref{eq:l-pre}) is calculated by the observation of each agent, Predict-Future is a local-level characteristic of LTS-CG. Employing the message-passing mechanism of GNNs \cite{duan2024layerdiverse}, it enables agents to predict future observations based on their own current observations and the passed information from neighbouring agents.

% We input a $T$-step of observation ${\{O^{i}_{t+1:t+T}\}_{i=1}^{n}}$ into Eq.\ref{eq:gru} to product $H_{t+T}$ as the summary of the input. Then, with the $H_{t+T}$, Eq.\ref{eq:gru} is iteratively used $T$ times to continue the recurrence of the hidden state for forecasting how much the current $T$-step observation will change denoted as ${\{\Delta O^{i}_{t+1:t+T}\}_{i=1}^{n}}$. The hidden state $H_{t+T}$ simultaneously serves as the output ${\{\Delta O^{i}_{t^{\prime}}\}_{i=1}^{n}}$ and the input to the next step for each iteration. The loss function for Predict-Future Decoder is now defined as:
% \begin{equation}
% L_{obs}= \sum_{i}^{n} \sum_{t^{\prime}=1}^{T}\left\| \left( O^{i}_{t+t^{\prime}}+ \Delta O^{i}_{t+t^{\prime}} \right) - O^{i}_{t+T+t^{\prime}}\right\|_{2}
% \end{equation}
% This decoder enables agents to anticipate the dynamics and make informed decisions based on the expected outcomes.
\noindent \textbf{Infer-Present} is
designed to assist every partially observed agent in gaining the ability to grasp the entire environmental context and deduce the current state with the information provided by the graph. Given the current observation ${\{o^{i}_t\}_{i=1}^{n}}$, we first generate the observation embeddings matrix $E_t=[e_{t}^{1\top},...,e_{t}^{n\top}]$ using the ongoing observation extractor $e_t^i= f_{obs}(o_t^{i})$, where $f_{obs}$ is a MLPs. Then we adopt an attention mechanism to dynamically calculate the edge weight between every pair of agents resulting in the attention edge-weight matrix, defined as
\begin{equation}
    \mu_t^{ij}= \frac{\text{exp}(e_{t}^{j\top}W_{a}e_{t}^{i})}{\sum_{k\in \mathcal{N}_{i}}\text{exp}(e_{t}^{k\top}W_{a}e_{t}^i)}, \quad C_t^{ij}= \mu_t^{ij},
\label{eq:attention}
\end{equation}
where $\mathcal{N}_{i}$ represents the neighbors of agent $i$ in the graph and $W_{a}$ is trainable parameter of attention mechanism. The weighted-agent-pair matrix is updated as $A_t^{\prime} = C_{t}A$, and the graph convolution \cite{DBLP:conf/iclr/KipfW17} is performed using the following equation 
 \begin{equation}
 \label{eq:GCN_MATIX}
    H_{t}^{l} = ReLU \Big( \hat{A}_t H_{t}^{(l-1)} W^{(l-1)} \Big),
\end{equation}
where $l$ is the index of GNN layers, $\hat{A}_t = \tilde{D}^{-\frac{1}{2}}A^{\prime}_{t}\tilde{D}^{-\frac{1}{2}}$, $\tilde{D}_{ii} = \sum_{j} A^{\prime}_{t}[i,j]$, and $H_{t}^{0}=E_{t}$ . The current sparse graph $A_t^{\prime}$ not only encapsulates historical information within its structure but also captures the ongoing agent relationships through the edge weights. The message-passing mechanism of the GNN in Eq.(\ref{eq:GCN_MATIX}) enables agents to exchange their knowledge effectively at every time step. The current feature of the entire graph at the $t$-step is defined as
\begin{equation}
\label{GCN_readout}
   g_t = \text{READOUT}(\sum_{i}^N H_{t}[i,:]),
\end{equation}
where $\text{READOUT}(\cdot)$ is an average function aggregating all the agents' information to obtain the entire graph feature.  The Infer-Present is achieved by 
\begin{equation}
\mathcal{L}_{inf}=  \sum_{t=1}^{T}\left\| g_t - s_t\right\|_{2},
\end{equation}
where $s_t$ denotes the actual state of the environment at the $t$ step. Infer-Present is a global-level characteristic of LTS-CG that utilizes graph convolution to facilitate a seamless exchange of observations among agents, allowing the entire graph (comprising all agents/nodes and their relationships/edges) to represent the current state of the environment collectively.

With the above two characters, the generalized loss function for the graph
learning Eq.(\ref{eq:learnMatrix}) now can be formalized as
\begin{equation}
    \mathcal{L}\left(A, w, \mathcal{O}_T \right) = \mathcal{L}_g =\mathcal{L}_{pre} + \mathcal{L}_{inf}.
\label{eq: graph-loss}
\end{equation}

\subsection{Cooperative MARL with LTS-CG}
\textcolor{black}{In our LTS-CG design, graph inference and multi-agent policy learning are integrated for efficient end-to-end training. The \textbf{Inter-Agent Sparse Graph Learning} module follows an encoder-decoder framework: the encoder generates the sparse graph structure, while the decoder—guided by the two graph loss functions mentioned earlier—refines the encoder’s weights.}

\textcolor{black}{At the start of training, the buffer stores a fully connected inter-agent graph, allowing agents to cooperate and make informed decisions from the outset. As training progresses, our method learns a temporally sparse graph, which is stored in the buffer and reused in subsequent training iterations, thereby accelerating the learning process. During testing, only the encoder is used to generate the graph structure. The resulting graph is then fed into an attention-based graph convolution part of \textbf{Cooperative MARL} module, dynamically adjusting edge weights at each time step. This temporal sparse graph ensures that agents always have access to the most up-to-date information for effective decision-making and coordination throughout the training process.}

Leveraging the learned graph $A$ at every time step and following the Eq.(\ref{eq:attention}), the current observation ${\{o^{i}_t\}_{i=1}^{n}}$ are used to compute the edge weights in $A$. These edge weights determine the importance of cooperating with neighbouring agents. Consequently, we obtain the latent temporal sparse coordination graph, encompassing historical information within its structure and ongoing agent relationships through its edge weights. The exchanged knowledge $m_i = H_{t}^{l}[i,:]$ between agents is then shared on this graph. Using Eq.(\ref{eq:GCN_MATIX}), what information should be exchanged is calculated during cooperation. This process enhances the agents' perception, prediction, and decision-making capabilities.  With this knowledge, the local action-value function is defined as $Q_i(\tau_{i},a_i,m_i)$. To keep the balance of computational efficiency with effective agent interaction and complex decision-making, we build our algorithm on top of the QMIX \cite{DBLP:conf/icml/RashidSWFFW18} to integrate all the individual Q values. The total-action value is monotonic in the per-agent values, which is formulated as
\begin{equation}
        \underset{\mathbf{a}}{\operatorname{argmax}} \quad Q_{t o t}(\boldsymbol{\tau}, \mathbf{a})=\left(\begin{array}{c}\operatorname{argmax}_{a_1} Q_1\left(\tau_1, a_1,m_1\right) \\ \vdots \\ \operatorname{argmax}_{a_n} Q_n\left(\tau_n, a_n, m_n\right)\end{array}\right).
\end{equation} 
The entire framework is trained by minimizing the loss function 
\begin{equation}
 \mathcal{L}(\boldsymbol{\theta})=\mathcal{L}_{TD}(\boldsymbol{\theta}^{-})+\lambda \mathcal{L}_g\left(\boldsymbol{\theta}_g\right), 
\label{eq:final-loss}
\end{equation}
where $\boldsymbol{\theta}$ includes all parameters in the model, $\mathcal{L}_g$ represents the graph loss from Eq. (\ref{eq: graph-loss}) and  $\lambda$ is the weight of graph loss. The TD loss $\mathcal{L}_{TD}(\boldsymbol{\theta}^{-})$ in Eq. (\ref{eq:final-loss}) is defined as 
\begin{equation}
\mathcal{L}_{TD}(\boldsymbol{\theta}^{-})\!=\!\left[r\!+\!\gamma \max _{\boldsymbol{a}^{\prime}} Q_{tot}\left(s^{\prime}, \boldsymbol{a}^{\prime} ; \boldsymbol{\theta}^{\prime}\right)\!-\!Q_{tot}(s, \boldsymbol{a} ; \boldsymbol{\theta}^{-})\right]^2,
\label{eq:loss}
\end{equation}
where $\boldsymbol{\theta}^{\prime}$ denotes the parameters of a periodically updated target network, as commonly employed in DQN. By training with the Eq.~(\ref{eq:final-loss}), our method enables simultaneous graph inference and multi-agent policy learning, facilitating efficient end-to-end training using standard policy optimization methods.

% These two decoders work in tandem to optimize graph learning in MARL by leveraging the learned graphs and incorporating comprehensive environmental information. They enhance the agents' perception, prediction, and decision-making capabilities, leading to improved performance and effectiveness in complex multi-agent scenarios.

\begin{figure*}[th]
\centering
\includegraphics[width=\textwidth]{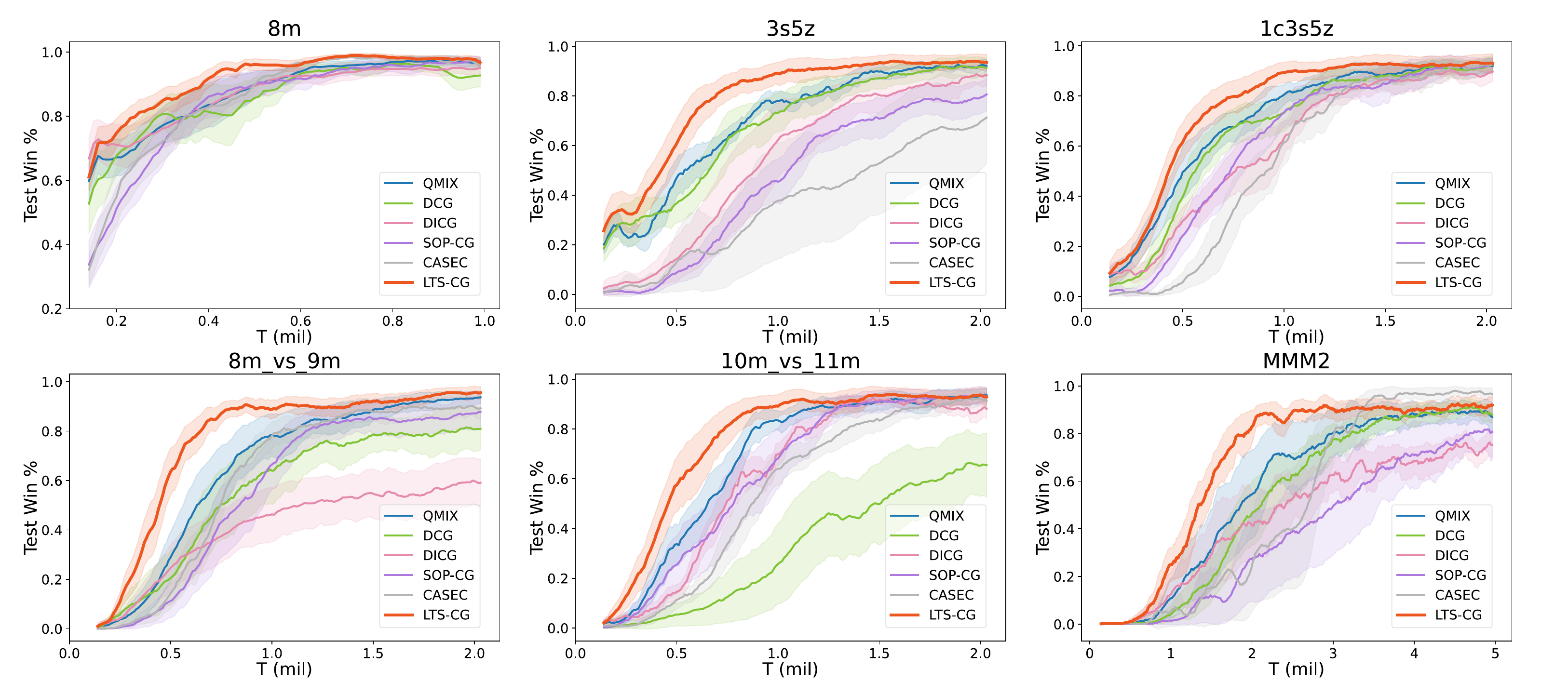} % Reduce the figure size so that it is slightly narrower than the column
\caption{Performance of our method and baselines on six maps of the StarCraft II benchmark \cite{DBLP:conf/atal/SamvelyanRWFNRH19}. The Y-axis is the test winning rate of the game. The X-axis is the training steps. }
\label{fig:Result}
\end{figure*}

\section{Experiments}
\label{sec:exp}
In this section, we design experiments to answer the following questions: (1) How does LTS-CG compare in performance with \textcolor{black}{graph-based} methods on complex cooperative multi-agent tasks? (See: \ref{sec:Graph-based}) \textcolor{black}{(2) How does LTS-CG compare in performance with non-graph methods? (See: \ref{sec:non-graph-based}) (3) How does LTS-CG perform across a variety of scenarios? (See: \ref{sec:TagGather})} (4) Is the utilization of trajectories for learning the coordination graph more effective than relying on one-step observations? (See: \ref{sec:traj-vs-one}) \textcolor{black}{(5) Is sampling from the Attention Matrix necessary? (See: \ref{sec:sparse matrix})} (6) Does having the Predict-Future and Infer-Present characteristics improve the performance of LTS-CG? (See: \ref{sec:graph strategy}) (7) What are the effects of varying the weights for $\mathcal{L}_g$ on the experimental outcomes? (See: \ref{sec:weight-loss})

To answer the above questions, \textcolor{black}{our experiments involve the following three environments:}
\begin{itemize}
    \item \textbf{StarCraft II benchmark} (SMAC) \cite{DBLP:conf/atal/SamvelyanRWFNRH19}: consists of different maps with varying numbers of agents. Our experiments included scenarios with a minimum of eight agents, comprising both homogeneous and heterogeneous agent setups. All the experiments are carried out with \textit{difficulty=}7.
    \item \textcolor{black}{\textbf{Tag} (MPE) \cite{DBLP:conf/nips/LoweWTHAM17}: is a task based on the particle world environment. In this scenario, a group of agents chases several adversaries on a map containing three randomly generated obstacles. The agents receive a global reward for each collision with an adversary. The adversaries move faster, making it crucial for the agents to collaborate effectively to surround them. We tested the common setup of 10 agents chasing 3 adversaries, and we further extended this to 20 agents chasing 5 adversaries to evaluate the scalability of different methods.}
    \item \textcolor{black}{\textbf{Gather}: is an extension of the Climb Game \cite{DBLP:journals/jmlr/WeiL16}. In the original Climb Game, each agent has three possible actions, $A= \{a_0, a_1, a_2\}$. Action $a_0$ yields no reward unless all agents choose it, at which point it provides a high reward. The other two actions are sub-optimal but can yield positive rewards without requiring perfect coordination. We followed the setup from the MULTI-AGENT COORDINATION BENCHMARK (MACO) \cite{DBLP:conf/iclr/00010DY0Z22} for our experiments. }
\end{itemize}

We employ distinct 2-layer GNNs as specified in Eq.(\ref{eq:GCN_MATIX}) to facilitate the acquisition of the Infer-present characteristic and to compute the knowledge exchanged during agents' cooperation. The graph loss $\lambda$, the character-balance weight $b$ and $c$ in Eq~(\ref{eq:final-loss}) are set to 1. 
% The hyperparameters used for MARL in StarCraft II are given in Tab.~\ref{tab:Hyper-SMAC}. 
Experiences are stored in a first-in-first-out (FIFO) replay buffer during the training phase, and all settings are repeated with five random seeds for consistency. The experiments are finished with Intel Xeon Gold 6226R CPUs and NVIDIA Quadro RTX 8000 GPUs (48 GB) GPU. The software that we use for experiments is Python 3.7.13, PyTorch 1.13.1, PyYAML 6.0, numpy 1.21.5 and CUDA 11.6. \textcolor{black}{More experimental details and our implementation can be found at \textit{https://github.com/Wei9711/LTSCG}}

\subsection{Performance Comparison on StarCraft II}

\subsubsection{\textcolor{black}{Comparison with Graph-based Methods}}
\label{sec:Graph-based}
We utilize several state-of-the-art baseline algorithms for our experiments. 
% Each method's graph type, edge representation, and group utilization are summarised in Tab.~\ref{tab:method}.  
Below, we provide a brief introduction of each method and the detailed settings we used:
\begin{itemize}
    \item \textbf{QMIX} \footnote{https://github.com/oxwhirl/pymarl} \cite{DBLP:conf/icml/RashidSWFFW18} is effective but without cooperation between agents. We adopt the configuration specified in the StarCraft Multi-Agent Challenge \cite{DBLP:conf/atal/SamvelyanRWFNRH19} for the QMIX algorithm.
    \item \textbf{DCG} \footnote{https://github.com/wendelinboehmer/dcg} \cite{DBLP:conf/icml/BoehmerKW20} directly links all the agents to get an unweighted fully connected graph. The graph is used to calculate the action-pair values function. For DCG, we employ a low-rank payoff approximation with $K=1$ (as described in Eq.(5) of the original paper) and incorporate privileged information through the action representation learning technique. This corresponds to the \textit{DCG-S (rank 1)} setting outlined in the original paper.

    \item \textbf{DICG} \footnote{https://github.com/sisl/DICG} \cite{DBLP:conf/atal/LiGMAK21}  uses attention mechanisms to calculate weighted fully connected graph. The graph is used to pass information between agents.  We utilize the DICG algorithm in the context of the centralised training centralised execution (CTCE) paradigm. This approach involves using QMIX as the base policy learning framework. The graph learning procedure strictly follows the DICG methodology.

    \item \textbf{SOP-CG} \footnote{https://github.com/yanQval/SOP-CG} \cite{DBLP:conf/icml/YangDRW0Z22} selects sparse graphs from a pre-calculated candidate set. In line with the original paper, we adopt the \textit{tree organization} $\mathcal{G}_T$ for SOP-CG. In this configuration, the agents are organized in a tree structure with $n-1$ edges, ensuring that all agents form a connected component.

    \item \textbf{CASEC} \footnote{https://github.com/TonghanWang/CASEC-MACO-benchmark}\cite{DBLP:conf/iclr/00010DY0Z22} drops some edges on the weighted fully connected graph according to the variance payoff function.  We employ the \textit{construction\_q\_var} (Eq.(4) in the paper) and \textit{q\_var\_loss} (Eq. 8 in the paper) strategies described in the original paper. The weight of the sparseness loss term is set to $\lambda_{sparse}=0.3$ in our experiments.

\end{itemize}

\begin{figure}[t]
\centering
\includegraphics[width=\columnwidth]{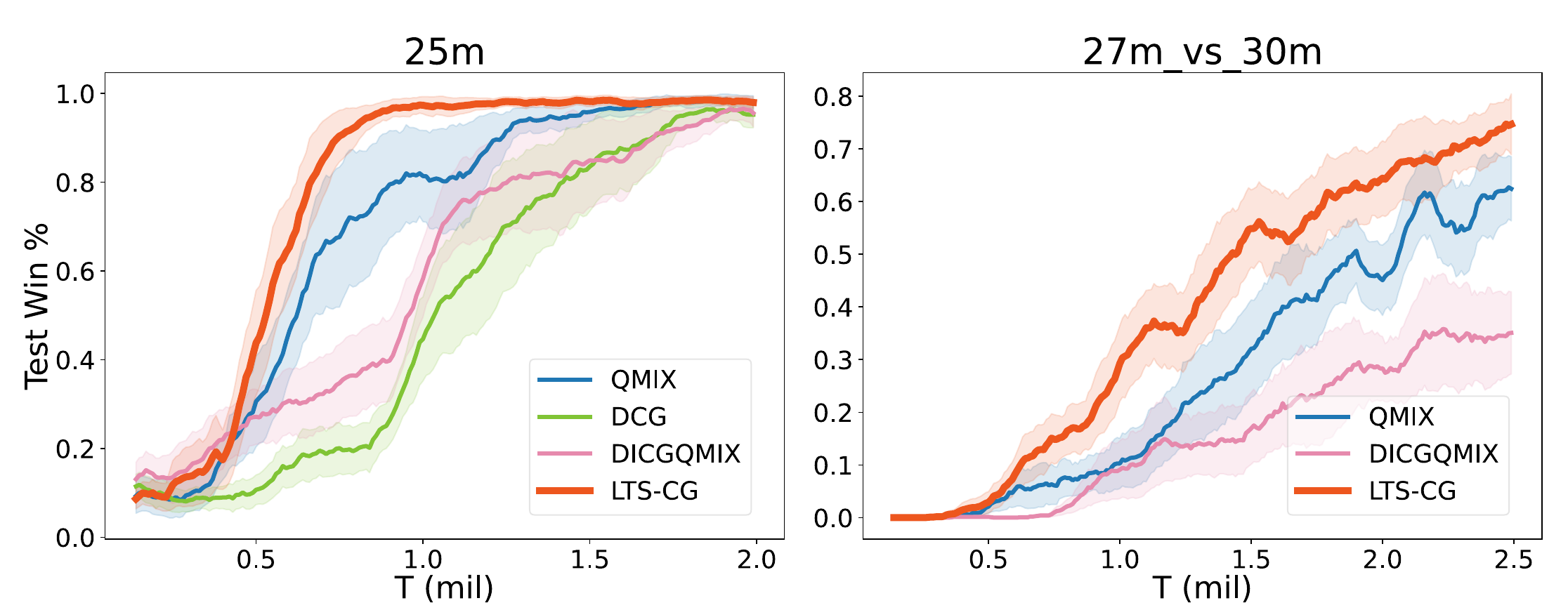}
\caption{Performance comparison on the \textit{25m} and \textit{27m\_vs\_30m} maps. \textcolor{black}{Due to the high computational complexity, SOP-CG and DCG could not complete 2 million steps within a week, and CASEC exceeded the 48 GB GPU memory limit.}}
\label{fig:27m}
\end{figure}

\noindent \textbf{Results}: 
Fig. \ref{fig:Result} presents the results of our method compared to the performance of other algorithms on six different maps. The experimental results clearly demonstrate the superiority of our approach LTS-CG across all scenarios (shown in black). Firstly, our method exhibited faster convergence than the compared methods on all six maps in the early stages of training (below 0.6 mil for \textit{8m}, 2 mil for \textit{MMM2}, and 1 mil for other maps). This indicates that our approach enables the agents to quickly learn effective cooperative strategies and achieve high-performance levels. Moreover, our method demonstrated a smaller standard deviation in performance compared to the other methods, such as CASEC in \textit{3s5z}, DICG in \textit{8m\_vs\_9m} and DCG in \textit{10m\_vs\_11m}. The reduced variability suggests that our approach consistently produces reliable and stable cooperative behaviours, resulting in more predictable and robust performance across different maps. Notably, our method achieved consistent and competitive performance across all six maps. This indicates that our approach generalizes well and is capable of adapting to various environmental conditions and agent configurations. The ability to achieve good results consistently is essential for real-world applications of multi-agent systems.

Comparing our method to two SOTA approaches, SOP-CG and CASEC, which aim to learn sparse graphs for MARL, we observed interesting patterns in their performance on specific maps. In the \textit{3s5z}, \textit{1c3s5z}, and \textit{10m\_vs\_11m} maps, SOP-CG outperformed CASEC. However, in the \textit{8m\_vs\_9m} and \textit{MMM2} maps, CASEC exhibited superior performance compared to SOP-CG. The varying performance of SOP-CG and CASEC indicates the importance of learning the meaningful graph based on the environment and agent setup, which further highlights the advantages of our approach in achieving constant and competitive performance across diverse scenarios.

\begin{figure}[t]
\centering
\includegraphics[width=\columnwidth]{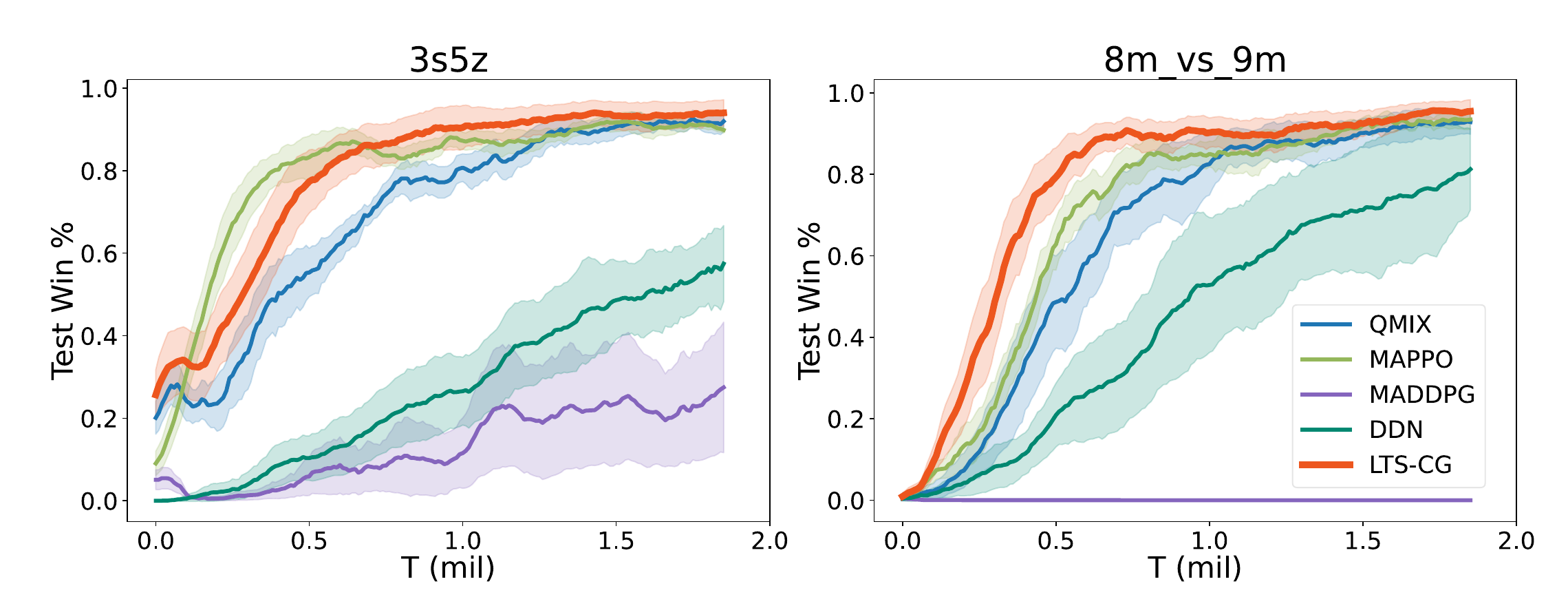}
\caption{\textcolor{black}{Performance comparison of non-graph-based methods on \textit{3s5z} and \textit{8m\_vs\_9m}.}}
\label{fig:ADD-SOTA}
\end{figure}

\noindent \textbf{Large maps.}  We further investigated the performance of the proposed method on larger maps: \textit{25m} and \textit{27m\_vs\_30m}, which are designed to test the scalability and efficiency of the algorithms under high computational complexity. Due to the high computational demands in representing action-pairs, two SOTA approaches, SOP-CG and CASEC, could not complete the experiments on both maps, and DCG could not finish the experiment on the \textit{27\_vs\_30m} map,  which is indicative of their computational limitations in this context. In Fig. \ref{fig:27m}, the results of our proposed method on these two maps were presented. Our approach demonstrated promising performance compared to the other methods, even in these challenging and computationally intensive scenarios. Notably, the QMIX algorithm (shown in blue), which operates without explicit cooperation mechanisms or coordination graphs, surprisingly outperforms DCG and DICG (shown in light green and pink, respectively), which are graph-based learning algorithms. This result indicates that while the graph-based approaches are designed to foster coordination among agents, the lack of a well-constructed coordination graph can be detrimental, potentially hindering the policy learning process.

In summary, the experiments suggest that graph-based coordination in multi-agent settings must be carefully crafted to ensure that it is conducive to the learning environment. The results highlight the necessity for well-designed graph structures that enhance rather than impede policy learning, as evidenced by the success of LTS-CG in complex scenarios where other graph-based methods struggle.

% \begin{table}[!t]

% \begin{center}
% \begin{tabular}{ccc}
% \toprule
%        & $10k$ steps time $(s)$ & GPU Memory (MB) \\
% \midrule
% QMIX   & $121$    & $1251$     \\
% DCG   & $421$    & $1563$     \\
% DICG   & $194$    & $1553$     \\
% SOP-CG & $375$     &   $2649$   \\
% CASEC &  $297$    &  $7805$    \\
% LTS-CG &  $299$    &  $2749$    \\
% \bottomrule
% \end{tabular}
% \end{center}
% \caption{8m\_vs\_9m}
% \label{tab:8m}
% \end{table}

% \begin{figure}[h]
% \centering
% \includegraphics[width=\columnwidth]{TNNLS/fig/LTSCG-25m.pdf} % Reduce the figure size so that it is slightly narrower than the column.
% \caption{Performance comparison on the \textit{25m} and \textit{27m\_vs\_30m} maps. Due to the high computational complexity, SOP-CG and CASEC could not complete the experiments on both maps. DCG could not finish the experiment on the later map.}
% \label{fig:27m}
% \end{figure}

\begin{figure*}[th]
\centering
\includegraphics[width=\textwidth]{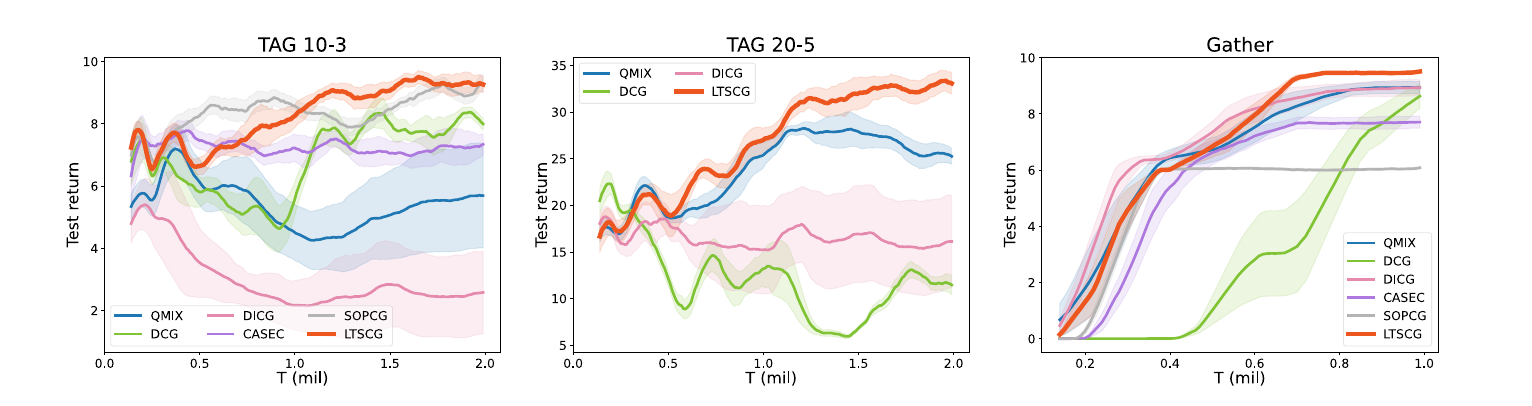} % Reduce the figure size so that it is slightly narrower than the column
\caption{\textcolor{black}{Performance comparison of different methods on the TAG  and Gather scenarios. The TAG scenario involves agents chasing adversaries on a map with obstacles, showing results for 10 agents and 3 adversaries and 20 agents and 5 adversaries The Gather scenario is an extension of the Climb Game where precise coordination yields higher rewards. }}
\label{fig:ADD-Env}
\end{figure*}

% \begin{figure}[t]
% \centering
% \includegraphics[width=\columnwidth]{fig/LTSCG-Atten.pdf} % Reduce the figure size so that it is slightly narrower than the column.
% \caption{Performance comparison on the \textit{3s5z} and \textit{10m\_vs\_11m}. \textit{OneStepObs-c} and \textit{OneStepObs-s} utilize a one-step observation to generate a fully connected graph and a sparse graph separately. \textit{LTS-CG}$(w/o \mathcal{L}_g)$ investigates the effectiveness of using trajectories by omitting the Predict-Future and Infer-Present characteristics.}
% \label{fig:atten}
% \end{figure}

% \begin{figure}[t]
% \centering
% \includegraphics[width=\columnwidth]{fig/LTSCG-loss.pdf} % Reduce the figure size so that it is slightly narrower than the column.
% \caption{Evaluate the effectiveness of the different latent temporal sparse graph learning strategies.}
% \label{fig:ablation-loss}
% \end{figure}
\begin{figure}[t]
\centering
\includegraphics[width=\columnwidth]{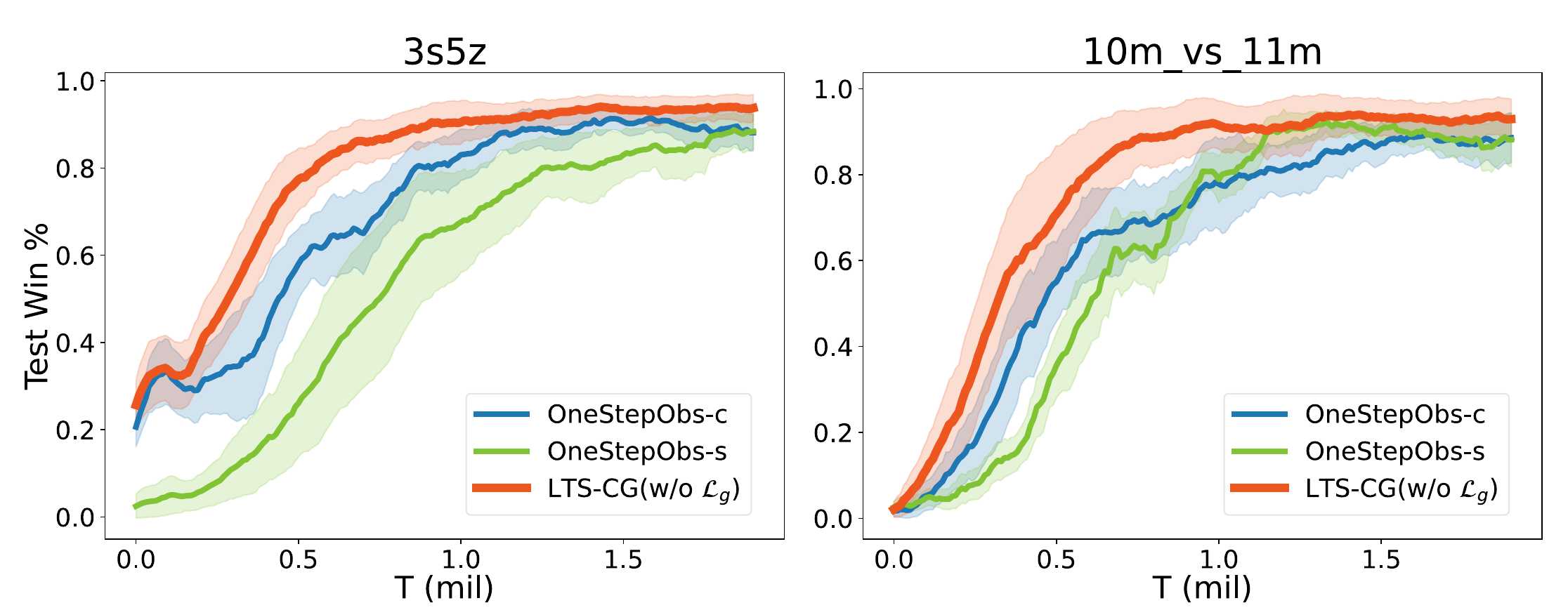}
\caption{Performance comparison on two maps to evaluate whether utilizing trajectories is more effective than relying solely on one-step observations. } %\textit{OneStepObs-c} and \textit{OneStepObs-s} utilize a one-step observation to generate a fully connected graph and a sparse graph separately. \textit{LTS-CG}$(w/o \mathcal{L}_g)$ investigates the effectiveness of using trajectories by omitting the Predict-Future and Infer-Present characteristics.}
\label{fig:atten}
\end{figure}

\subsubsection{\textcolor{black}{Comparison with Non-graph-based Methods}}
\label{sec:non-graph-based}
\textcolor{black}{
In this subsection, we further include several non-graph-based methods for comparison with our proposed method, as these are widely accepted as benchmarks and are frequently used to assess the performance of new algorithms in the SMAC environment:
\begin{itemize}
    \item \textbf{DDN} \footnote{https://github.com/j3soon/dfac}\cite{DBLP:conf/icml/SunLL21} uses distributional reinforcement learning to factorize the value function by modelling utility functions as random variables and applying a quantile mixture. We standardized the agent dimensions to 64 instead of the original 256 for consistency across methods.
    \item \textbf{MAPPO} \footnote{https://github.com/marlbenchmark/on-policy} \cite{DBLP:conf/nips/YuVVGWBW22} is a widely used method for cooperative multi-agent reinforcement learning, which has been shown to perform well in both continuous and discrete action spaces.
    \item \textbf{MADDPG} \footnote{https://github.com/uoe-agents/epymarl}\cite{DBLP:conf/nips/LoweWTHAM17} is designed for mixed cooperative-competitive environments and leverages centralized training with decentralized execution.
\end{itemize}
}
\noindent 
\textcolor{black}{ \textbf{Results:} Fig. \ref{fig:ADD-SOTA} shows the test win rates of the compared methods across different training steps. Our proposed LTS-CG method demonstrates competitive and consistent performance on the selected maps. On the \textit{3s5z} map, while MAPPO initially converges faster, LTS-CG surpasses it with a slightly higher win rate post-convergence. On the \textit{8m\_vs\_9m} map, LTS-CG outperforms MAPPO.
}
\textcolor{black}{ 
DDN, standardized to a 64-dimensional RNN agent in our experiments, fails to achieve competitive results compared to LTS-CG on both maps. Similarly, MADDPG struggles, with its win rate remaining below 50$\%$ after 2 million training steps, indicating limited effectiveness.
}

\begin{figure}[t]
\centering
\includegraphics[width=\columnwidth]{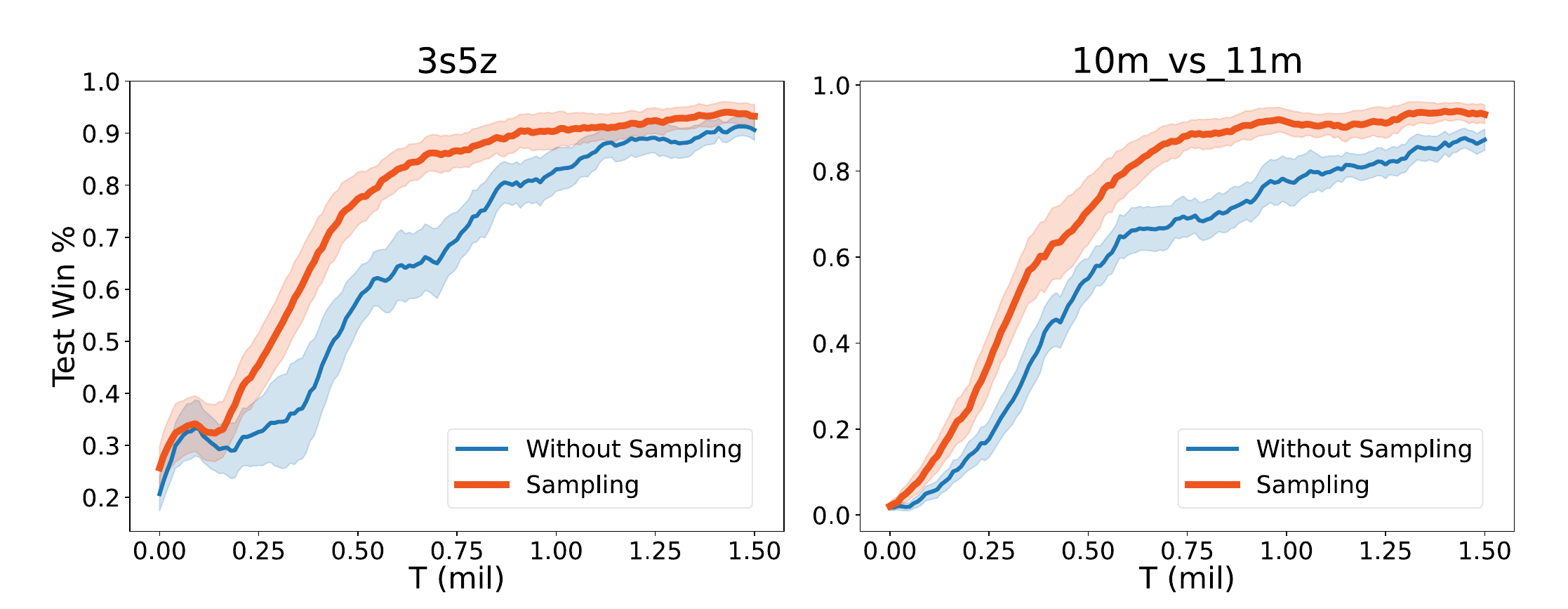}
\caption{\textcolor{black}{Performance comparison on two maps to evaluate whether sampling the graph from the attention matrix is more effective than not sampling.}}
\label{fig:Abla-Atten}
\end{figure}

\subsection{\textcolor{black}{Performance Comparison on Tag and Gather}}
\label{sec:TagGather}

\textcolor{black}{We further evaluated the performance of different methods on the TAG and Gather scenarios. Fig. \ref{fig:ADD-Env} shows the results for 10 agents and 3 adversaries and the results for 20 agents and 5 adversaries. The Gather scenario is an extended version of the Climb Game, where precise coordination is essential for achieving higher rewards.}

\textcolor{black}{Our proposed method, LTS-CG, consistently demonstrates competitive performance across both scenarios. In the TAG scenario, LTS-CG scales efficiently from 10 to 20 agents, maintaining robust performance. Notably, when the number of agents increases to 20, two state-of-the-art methods, SOP-CG and CASEC, fail to complete training within 7 days under the same experimental conditions. The performance of DCG drops significantly when scaling from 10 to 20 agents. 
In the Gather scenario, LTS-CG exhibits the best test return after 0.8 million steps and demonstrates a promising convergence speed compared to DICG.
These results illustrate the scalability, robustness, and efficiency of LTS-CG across a variety of environments and agent numbers, proving its capability to handle complex multi-agent scenarios.}

\begin{figure*}[t]
\centering
\includegraphics[width=\textwidth]{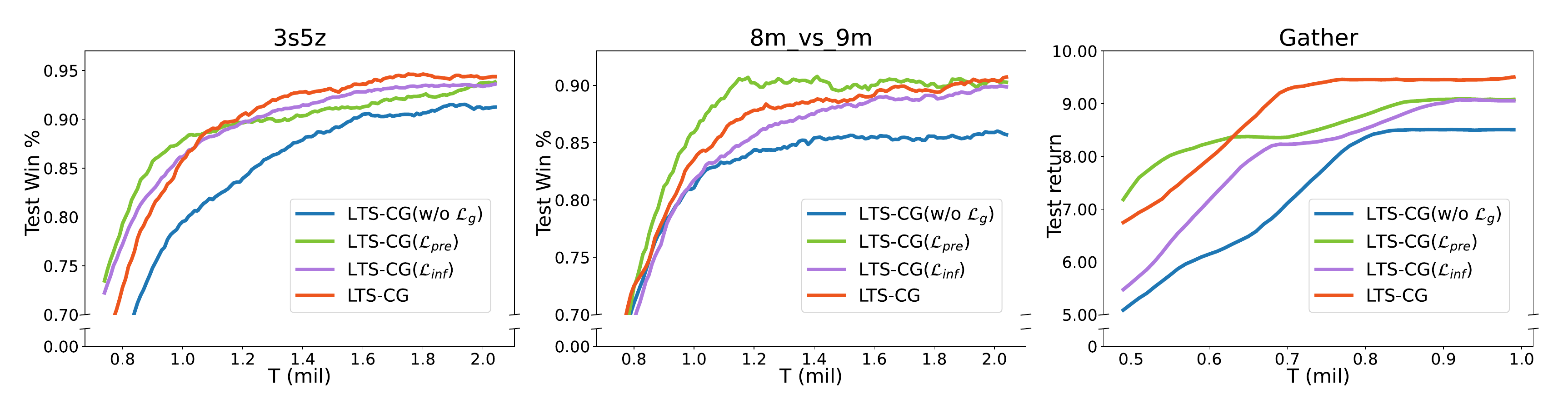}     
\caption{\textcolor{black}{Evaluate the effectiveness of the different latent temporal sparse graph learning strategies on SMAC and Gather. }}%\textit{LTS-CG}$(w/o \mathcal{L}g)$ excludes both Predict-Future and Infer-Present characteristics. \textit{LTS-CG}$(\mathcal{L}{pre})$ and \textit{LTS-CG}$(\mathcal{L}_{inf})$ only incorporate the Predict-Future or Infer-Present characteristic into the learning process separately.}}
\label{fig:ablation-loss}
\end{figure*}

\subsection{Ablation Study}
\subsubsection{Trajectory Graph Learning vs One-Step Observations}
\label{sec:traj-vs-one}
We examined the effect of graph generation methods on MARL performance in the \textit{3s5z} and \textit{10m\_vs\_11m} scenarios. We considered three settings: 
\begin{itemize}
    \item \textit{OneStepObs-c} generates a fully connected graph using one-step observations, akin to methods like DICG \cite{DBLP:conf/atal/LiGMAK21}. 
    \item \textit{OneStepObs-s} employs one-step observations to create a sparse graph, similar to G2ANet \cite{DBLP:conf/aaai/LiuWHHC020}. 
    \item \textit{LTS-CG}$(w/o \mathcal{L}_g)$ utilizes trajectories for graph generation while excluding Predict-Future and Infer-Present characteristics to solely assess the impact of trajectory-based learning.
    
\end{itemize}

As depicted in Fig. \ref{fig:atten}, \textit{LTS-CG}$(w/o \mathcal{L}_g)$ surpasses both \textit{OneStepObs-c} and \textit{OneStepObs-s} in win percentage over training iterations, demonstrating its superior performance in cooperative multi-agent settings. This finding underscores the significant benefit of trajectory-based graph generation in enhancing MARL performance, independent of other factors. The shaded areas in the figure represent the variance across multiple runs, with \textit{LTS-CG}$(w/o \mathcal{L}_g)$ not only achieving higher win rates but also exhibiting less variance, reflecting its consistent and reliable performance.

Furthermore, in Fig. \ref{fig:atten}, the comparison among \textit{LTS-CG}$(w/o \mathcal{L}_g)$, \textit{OneStepObs-c} (a method similar to DICG), and \textit{OneStepObs-s} (a method similar to G2ANet) shows that \textit{LTS-CG}$(w/o \mathcal{L}_g)$ demonstrates the most significant performance improvement in terms of win percentage across training iterations. This outcome highlights the advantages of using trajectory-based information for graph generation, even without relying on specialized characteristics like Predict-Future and Infer-Present.

The shaded regions in the graph represent the variance in win percentages over multiple runs, providing insights into the reliability of the methods. Notably, \textit{LTS-CG}$(w/o \mathcal{L}_g)$ achieves higher win rates and maintains tighter confidence intervals, suggesting a consistent performance advantage over the other methods. These experimental results provide strong support for the hypothesis that trajectory-based graph learning is more effective and robust than one-step observation-based methods, contributing significantly to the advancement of cooperative multi-agent learning techniques.

\subsubsection{\textcolor{black}{The Necessity of Sampling from Attention Matrix}}
\label{sec:sparse matrix}
\textcolor{black}{We further investigated whether sampling the graph from the attention matrix (sparse graph) is more effective than not sampling (dense graph). In the latter case, the attention matrix is directly used as the adjacency matrix, resulting in a fully connected graph. These studies were performed on the \textit{3s5z} and \textit{10m\_vs\_11m} maps. The results are presented in Fig. \ref{fig:Abla-Atten}.}

\textcolor{black}{Our method outperforms the fully connected graph approach, where the attention matrix is used directly. This suggests that relying attention-based matrix is insufficient for optimal performance. One potential reason for the lower performance  is the excessive exchange of messages between agents. While communication aims to enhance coordination, the large volume of irrelevant messages can overwhelm agents and distract them from making optimal decisions.}

\textcolor{black}{In contrast, our method's sparse graph mitigates this issue by restricting message passing to the most relevant agent pairs, reducing unnecessary information flow. This allows agents to focus on the most critical knowledge for decision-making. Furthermore, by treating the attention matrix as a distribution and sampling the graph from it, LTS-CG captures the inherent uncertainty in dynamic environments more effectively, leading to richer representations of agent cooperation and better adaptability over time.}

\subsubsection{Latent Temporal Sparse Graph Learning strategies} 
\label{sec:graph strategy}
We conducted an evaluation to assess the effectiveness of different strategies and examine the importance of the Predict-Future and Infer-Present characteristics in graph learning. Our investigation focused on the following settings: 
\begin{itemize} 
    \item \textit{LTS-CG}$(w/o \mathcal{L}_g)$ excludes both Predict-Future and Infer-Present characteristics. This setting implies that we do not further refine the learned graph structure after sampling.
    \item \textit{LTS-CG}$(\mathcal{L}_{pre})$ only incorporates the Predict-Future characteristic into the learning process.
    \item \textit{LTS-CG}$(\mathcal{L}_{inf})$ only incorporates the Infer-Present characteristic into the learning process.
    \item \textit{LTS-CG} with both Predict-Future and Infer-Present characteristics.
\end{itemize}
\textcolor{black}{The final performance is assessed on the \textit{8m\_vs\_9m}, \textit{3s5z} maps of SMAC and the Gather scenario.} The results are presented in Fig. \ref{fig:ablation-loss}. The ablation study revealed several important findings. Firstly, regardless of whether we include the Predict-Future, the Infer-Present, or both characteristics, the performance was consistently better than not having anyone. \textcolor{black}{This trend is consistent across different environments (SMAC and Gather), highlighting that each characteristic independently enhances agent coordination and overall performance.} Moreover, \textcolor{black}{since SMAC is known for its complexity and dynamic nature, our method's performance, incorporating both $\mathcal{L}_{pre}$ and $\mathcal{L}_{inf}$, exhibits some variation across different maps. This dynamic performance underscores that different maps may require different levels of emphasis on prediction and inference. To gain deeper insights into these dynamics, we extended our investigation to the Gather environment, which offers a different set of challenges and complexities. The results in this environment confirm the generalizability of our findings: $\mathcal{L}_{pre}$  is not redundant but plays a crucial role in improving performance when used in conjunction with $\mathcal{L}_{inf}$ . The synergy between these losses ensures that our method can effectively capture temporal dependencies and uncertainties in agent relationships, leading to superior outcomes in multi-agent coordination tasks.} This ablation study confirms the significance of two characteristics in LTS-CG for learning meaningful graphs to help agents cooperate.

% This ablation study confirms the significance of Predict-Future and Infer-Present characteristics in LTS-CG and their positive impact on the final performance. The decision of whether to include these characteristics depends on the agent types and the complexity of the task at hand.

% \begin{figure}[t]
% \centering
% \includegraphics[width=\columnwidth]{fig/LTSCG-lamda.pdf} % Reduce the figure size so that it is slightly narrower than the column.
% \caption{Evaluate the effect of the different weights of graph loss.}
% \label{fig:ablation-weight}
% \end{figure}
\begin{figure}[t]
\centering
\includegraphics[width=\columnwidth]{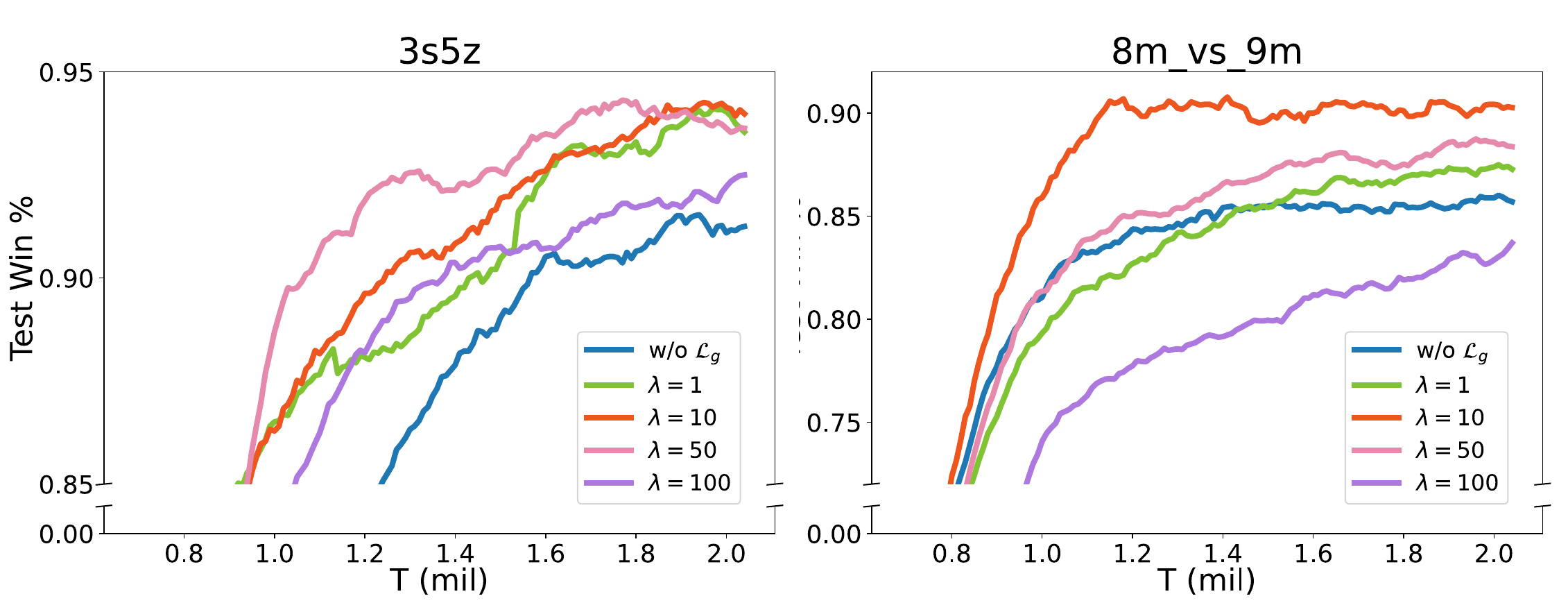}
\caption{Evaluate the effect of the different weights of $\mathcal{L}_g$.}
\label{fig:ablation-weight}
\end{figure}

\subsubsection{Weight of Graph Loss} 
\label{sec:weight-loss}
We tested the different weight of graph loss $\mathcal{L}_g$ on two maps, as shown in Fig. \ref{fig:ablation-weight}. $(w/o\text{ } \mathcal{L}_g)$ represents the scenario where the MARL training does not include the graph loss term, i.e., $\lambda=0$.  The results demonstrate the positive impact of incorporating $\mathcal{L}_g$ in MARL, as compared to the case without it. Specifically, when $\mathcal{L}_g =1,10,50$, the addition of $\mathcal{L}_g$ Consistently improves the performance of MARL on both maps. As the value of $\lambda$ increases, the final results during training on both maps first improve and then start to decline, which indicates that the weight $\lambda$ of the graph loss function has a noticeable influence on the final results. We present empirical evidence related to the parameter $\lambda$ here.

\begin{figure}[t]
\centering
\includegraphics[width=0.85\columnwidth]{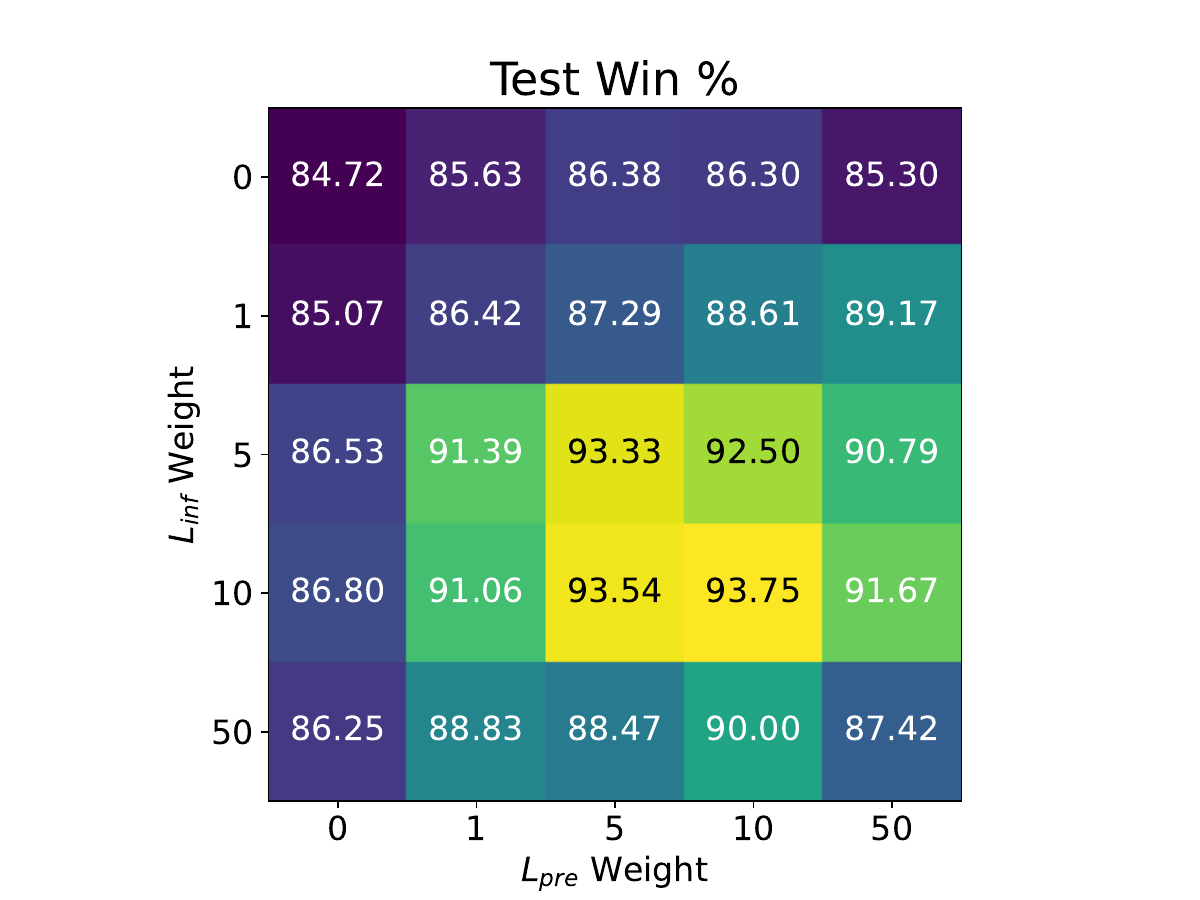}
\caption{\textcolor{black}{Heatmap illustrating the average test win rate on the \textit{8m\_vs\_9m} map in the SMAC environment under different configurations of the $\mathcal{L}_{pre}$ and $\mathcal{L}_{inf}$ weights. }}
\label{fig:LossMatrix}
\end{figure}

\textcolor{black}{Since $\mathcal{L}_g$ comprises two components, $\mathcal{L}_{pre}$ (predict-future) and $\mathcal{L}_{inf}$ (infer-present), we further investigated their individual contributions by testing different weight combinations $\{0, 1, 5, 10, 50\}$ for each, with 5 independent runs for each setting. The results, shown in Fig. \ref{fig:LossMatrix}, depict the average test win rate under various weight configurations. }

\textcolor{black}{
From this analysis, we observed the following key points: (1) Even a minimal inclusion of either $\mathcal{L}_{pre}$ or $\mathcal{L}_{inf}$ (i.e., weights greater than 0) consistently improves the test win rates compared to their exclusion. This finding validates the effectiveness of introducing the Predict-Future and Infer-Present mechanisms, leading to more meaningful graph construction and improved agent coordination. (2) As the weights of $\mathcal{L}_{pre}$ and $\mathcal{L}_{inf}$ increase from 0 to 10, the performance improves steadily. However, as the weights further increase to 50, the performance starts to degrade. This trend suggests that a moderate weighting of these losses is beneficial, while overly large weights may negatively affect performance by overemphasizing certain aspects of the learning process. (3) The best performance is achieved when both the weights of $\mathcal{L}_{pre}$ and $\mathcal{L}_{inf}$ are set to 10, as this configuration yields the highest test win rate, indicating an optimal balance between the two loss terms.}

Identifying the most appropriate $\lambda$ value for specific scenarios is a labour-intensive task that requires additional experimentation. It involves balancing leveraging the benefits of graph-based learning and avoiding potential overfitting or performance degradation due to excessive emphasis on the graph loss term. This process underscores the nuanced nature of parameter tuning in MARL and highlights the need for careful consideration when designing and optimizing such systems.

\begin{figure*}[t]
\centering
\includegraphics[width=\textwidth]{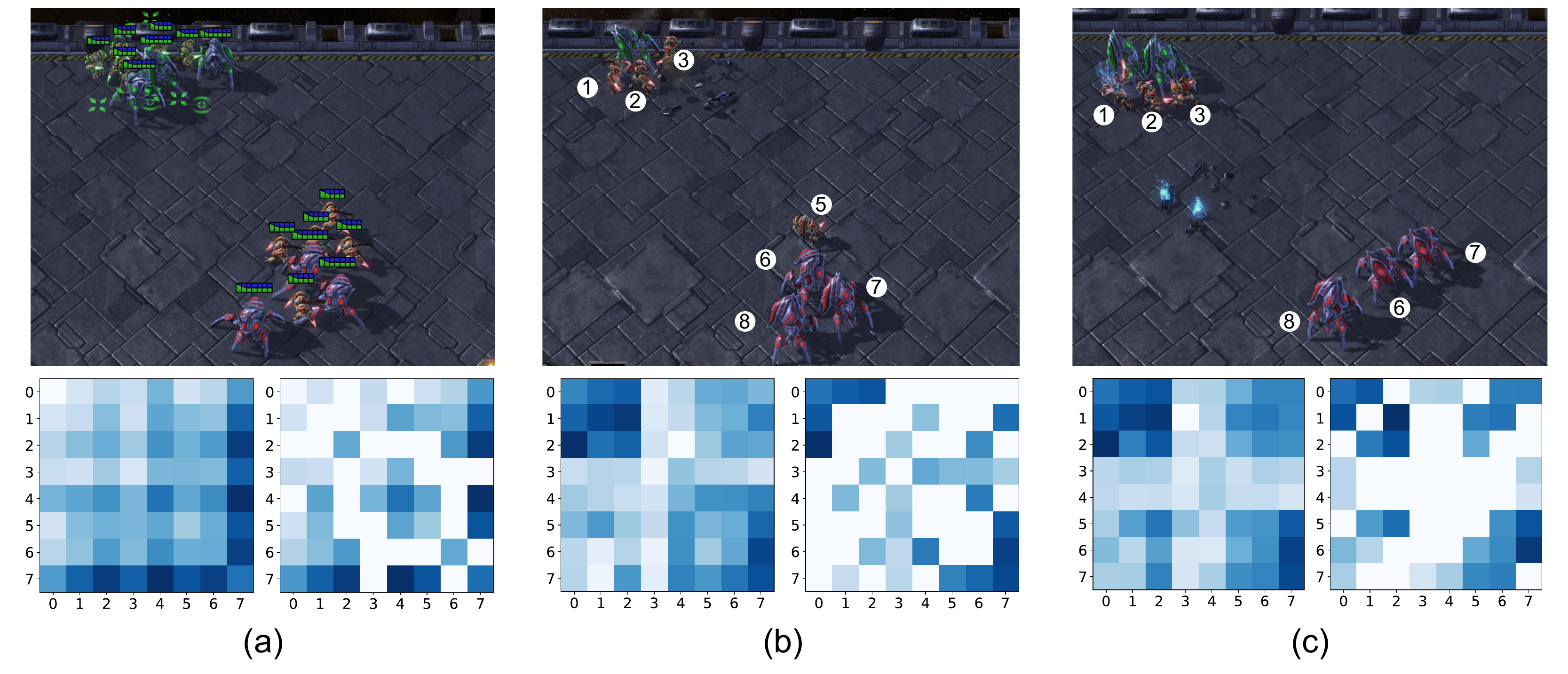} % Reduce the figure size so that it is slightly narrower than the column.
\caption{\textcolor{black}{A case study on the StarCraft II benchmark map \textit{3s5z}, featuring 3 Stalkers and 5 Zealots. The top row displays screenshots from the actual gameplay replay, while the bottom row illustrates the corresponding attention matrices and final sparse matrices.  }}
\label{fig:case-study}
\end{figure*}

\subsection{Case Study}
\textcolor{black}{
In this case study, we visualize the attention and sparse matrices alongside the actual game replay to demonstrate the interpretability of our model, shown in Fig.\ref{fig:case-study}. It highlights the most critical interactions among agents at different stages of the game: }
\begin{itemize}
    \item \textcolor{black}{\textbf{Scenario (a)}: At the beginning of the game, all agents exhibit high attention values towards each other (notably in the last column and row of the attention matrix), underscoring the importance of initial coordination. Even at this early stage, the sparse matrix begins reducing edges, refining communication to focus on key interactions. }
    \item \textcolor{black}{\textbf{Scenario (b)}: During the combat phase, where three Zealots are actively engaged while other agents remain on standby, the attention matrix reveals two distinct blocks that correspond to the two separate groups of agents. The sparse matrix emphasizes the importance of within-group communication over interactions between the groups at this point in the game. }
    \item \textcolor{black}{\textbf{Scenario (c)}: After the elimination of agents 4 and 5, the attention matrix shows reduced intensity in the rows and columns corresponding to these agents. The sparse matrix further prunes edges related to the eliminated agents, effectively modeling the decreased necessity for their participation in the communication network. }
\end{itemize}
\textcolor{black}{
These visualizations of the attention matrix help illustrate how LTS-CG dynamically captures the most relevant relationships among agents, contributing to a clearer understanding of the method.}

\begin{table}[t]
    \centering
    \begin{tabular}{ccccc}
        \toprule
        Method  & Graph type  & \begin{tabular}[c]{@{}c@{}}Sample \\ Edge\end{tabular}  & Data used & \begin{tabular}[c]{@{}c@{}}Graph Calculation \\ Time Complexity\end{tabular} \\
        \midrule
        QMIX     & $\times$  & $\times$   & $\times$ & $\times$\\
        DCG     & Complete & $\times$    & One-step & $O(A^2N^2)$\\
        DICG & Complete & $\times$     & One-step & $O(KN^2) $\\
        SOP-CG & Sparse  & $\times$     & One-step & $O(A^2N^2)$\\
        CASEC & Sparse  & $\times$     & One-step & $O(A^2N^2)$\\
        \textbf{LTS-CG} & \textbf{Sparse}  & \textbf{Yes}     & \textbf{Trajectories} & $O(TN^2)$\\
        \bottomrule
    \end{tabular}
    \caption{\textcolor{black}{Comparison of different experiment methods in terms of graph type, edge sampling, data used for learning the graph, and graph calculation time complexity.}}
    \label{tab:method}
\end{table}

\begin{table}[t]
\begin{center}
\begin{tabular}{cccc}
\toprule
Method       & 1k steps time (s)         & 1m steps time (h)        & GPU Memory \\
\midrule
QMIX   & $15.21 \pm 2.48$          & $2.7 \pm 0.41$           & 1.32 GB                \\
DCG    & $32.50 \pm 1.71$          & $11.25 \pm 1.47$         & 1.59 GB                \\
DICG   & $20.12 \pm 2.76$          & $7.32 \pm 1.69$          & 1.60 GB                \\
CASEC  & $28.50 \pm 4.65$          & $9.46 \pm 1.82$          & 6.38 GB                \\
SOP-CG & $24.22 \pm 5.68$          & $13.90 \pm 0.56$         & 2.45 GB                \\
LTS-CG & $20.76 \pm 2.47$          & $5.64 \pm 1.53$          & 3.17 GB                
\\
\bottomrule
\end{tabular}
\end{center}
\caption{\textcolor{black}{Running time and GPU consumption on \textit{8m}.}}
\label{tab:time_8m}
\end{table}

\begin{table}[t]
\begin{center}
\begin{tabular}{cccc}
\toprule
Method       & 1k steps time (s)         & 1m steps time (h)        & GPU Memory \\
\midrule
QMIX   & $20.13 \pm 3.59$          & $6.79 \pm 0.37$          & 1.50 GB                \\
DCG    & $33.57 \pm 4.65$          & $11.63 \pm 0.64$         & 2.37 GB                \\
DICG   & $20.74 \pm 4.37$          & $7.81 \pm 0.46$          & 2.03 GB                \\
CASEC  & $30.50 \pm 2.03$          & $10.12 \pm 0.51$         & 10.64 GB               \\
SOP-CG & $35.46 \pm 3.62$          & $19.46 \pm 0.80$         & 4.21 GB                \\
LTS-CG & $22.68 \pm 3.89$          & $8.84 \pm 0.49$          & 4.45 GB                
\\
\bottomrule
\end{tabular}
\end{center}
\caption{\textcolor{black}{Running time and GPU consumption on \textit{10m\_vs\_11m}.}}
\label{tab:time_10m}
\end{table}

\begin{table}[t]
\begin{center}
\begin{tabular}{cccc}
\toprule
 Method      & 1k steps time (s)         & 1m steps time (h)        & GPU Memory \\
\midrule
QMIX   & $26.28 \pm 4.58$          & $7.30 \pm 0.72$          & 1.93 GB                \\
DCG    & $43.67 \pm 5.73$          & $18.83 \pm 0.58$         & 13.35 GB               \\
DICG   & $27.79 \pm 6.65$          & $8.94 \pm 0.68$          & 3.39 GB                \\
CASEC  & /                         & /                        & Out of 48GB GPU        \\
SOP-CG & $516.04 \pm 10.76$        & More than 7 days         & 25.42 GB               \\
LTS-CG & $31.73 \pm 2.89$          & $10.37 \pm 0.53$         & 11.91 GB                
\\
\bottomrule
\end{tabular}
\end{center}
\caption{\textcolor{black}{Running time and GPU consumption on \textit{25m}.}}
\label{tab:time_25m}
\end{table}

\subsection{Discussion}
\label{sec:Dsicussion}
\textcolor{black}{This section first summarizes the compared methods in terms of graph type, edge sampling, data used for learning the graph, and graph calculation time complexity, as shown in Tab.\ref{tab:method}. Next, we highlight the key theoretical differences between our LTS-CG method and existing graph-based MARL approaches. We also relate this discussion to recent complementary work on group-aware coordination graphs \cite{DBLP:conf/ijcai/00030X24}, Bayesian ego-graph inference for networked MARL \cite{duan2025bayesian}, and bandwidth-constrained variational message encoding \cite{duan2026bandwidth}.}

\subsubsection{Graph as Coordination Graph vs. Graph for Message Passing}
\textcolor{black}{Existing methods like DCG \cite{DBLP:conf/icml/BoehmerKW20}, SOP-CG \cite{DBLP:conf/icml/YangDRW0Z22}, CASEC \cite{DBLP:conf/iclr/00010DY0Z22}, and GACG \cite{DBLP:conf/ijcai/00030X24} explicitly model coordination between agents using a coordination graph (CG). The graph represents action pair coordination, where the Q-function is factorized into utility functions $q^{i}$ and payoff functions $q^{ij}$ as Eq.~(\ref{eq:CG}).
This explicit coordination allows for direct evaluation of the coordination quality but incurs a high computational cost, $O(A^2N^2)$, due to the large number of action pairs required for the payoff functions $q^{ij}$. In contrast, methods like DICG \cite{DBLP:conf/atal/LiGMAK21} and LTS-CG infer implicit graphs that facilitate knowledge sharing during policy learning, bypassing direct action pair calculations; cooperative MARL with learned message encodings under bandwidth limits has also been studied \cite{duan2026bandwidth}. This approach significantly reduces computational complexity. For example, LTS-CG operates with a complexity of $O(TN^2)$, where $T$ is the trajectory length. }

\textcolor{black}{We present the detailed running time and GPU consumption for the compared methods on the \textit{8m} (Tab.~\ref{tab:time_8m}), \textit{10m\_vs\_11m} (Tab.~\ref{tab:time_10m}), and \textit{25m} (Tab.~\ref{tab:time_25m}) maps from SMAC. As shown in these tables, when the number of agents increases from 8 to 25, our LTS-CG method maintains acceptable computational resource consumption. In contrast, CASEC exceeds the 48GB GPU memory limit, and SOP-CG could not complete 1 million steps within one week on the \textit{25m} map. These findings demonstrate that LTS-CG scales efficiently with larger agent counts, offering a more practical solution for complex multi-agent scenarios compared to other graph-based methods, especially in terms of computational resources and runtime efficiency.}

\subsubsection{Sampling Graphs from Attention Matrix vs. Direct Use of Attention Matrix}
\textcolor{black}{
Unlike DICG \cite{DBLP:conf/atal/LiGMAK21}, which directly uses the attention matrix as the graph, LTS-CG introduces a novel sampling approach. By treating the attention matrix as a distribution and sampling graphs from it, LTS-CG effectively captures the uncertainty inherent in dynamic environments, in a spirit related to probabilistic graph views in networked MARL \cite{duan2025bayesian}. This sampling process allows for richer and more adaptable representations of agent cooperation, as the graphs evolve over time based on the sampled attention weights. Consequently, this approach enhances agent adaptability to changing environments, resulting in improved performance and coordination flexibility.}

\subsubsection{Trajectories vs. One-step Observation} 
\textcolor{black}{A key distinction in LTS-CG is the use of observation trajectories to generate the agent-pair probability matrix, rather than relying on single-step observations. We posit that observation trajectories provide a more comprehensive view of the temporal dynamics of agent interactions, leading to more accurate and meaningful graph representations. Empirical results in Sec. \ref{sec:traj-vs-one} support the validity of this assumption, demonstrating the effectiveness of trajectory-based graph construction.}

\subsubsection{Further Learning the Graph Characteristics}
\textcolor{black}{Many existing graph-based methods (e.g., DCG\cite{DBLP:conf/icml/BoehmerKW20}, DICG\cite{DBLP:conf/atal/LiGMAK21}) rely primarily on attention mechanisms to infer coordination graphs but often lack additional regularization techniques, which can lead to arbitrary or less informative edges.
LTS-CG addresses this issue by introducing two distinctive components: Predict-Future and Infer-Present. These mechanisms allow agents to anticipate future states and optimize their current coordination with limited data, respectively. By incorporating these features, LTS-CG constructs graphs that are not only spatially but also temporally optimized, leading to more meaningful and informed cooperation between agents. This enhanced graph learning, combined with regularization, results in significantly better collaboration and performance across complex, multi-agent environments.}

\section{Conclusions and Future directions}
\label{sec:conclusion}
This paper introduces LTS-CG, a novel approach for MARL that infers a latent temporal sparse graph to enable effective information exchange among agents. To efficiently infer the graph from past experiences, LTS-CG uses the agents' observation trajectories to generate the agent-pair probability matrix. Motivated by the idea that the meaningful graph should enrich agents’ comprehension of their peers and the environment, we further learn the graph to encode two essential
characteristics: Predict-Future and Infer-Present. The former is a local-level characteristic that gives agents valuable insights into the future environment, enhancing their decision-making capabilities in the current time step. The latter is a global-level one that enables partially observed agents to deduce the current state, promoting overall cooperation among agents. By having them, LTS-CG learns temporal graphs from historical and
real-time information, facilitating knowledge exchange during policy learning and effective collaboration. Graph learning and agent training occur simultaneously in an end-to-end manner. Experimental evaluations on the StarCraft II benchmark demonstrate the superior performance of our method over existing ones. 

%Extensive ablation studies confirm the efficacy of each individual component, underscoring the pivotal role of graph characterization and affirming LTS-CG's remarkable outcomes.

For future directions, it is imperative to extend the scope of graph learning beyond agent-pair relationships. Investigating higher-order relationships, such as group dynamics, while inferring cooperation graphs can deepen our understanding of cooperative behaviours among agents. Additionally, addressing the challenges posed by asynchronous scenarios is crucial. Developing techniques to effectively learn cooperation graphs in such scenarios will enhance the applicability and robustness of methods in real-world environments.

\section*{Acknowledgment}
This work is supported by the Australian Research Council under Australian Laureate Fellowships FL190100149 and Discovery Early Career Researcher Award DE200100245.
\bibliographystyle{IEEEtranN}
\bibliography{ltscg}

\end{document}